\documentclass[letterpaper, 10 pt, journal, twoside]{IEEEtran}

\IEEEoverridecommandlockouts                              

\usepackage{afterpage}

\usepackage{dblfloatfix} 

\usepackage{subfig}
\usepackage{tikz}
\usepackage{cite}
\usepackage{multirow,bigdelim}

\usepackage{amsmath}

\let\labelindent\relax
\usepackage{adjustbox}
\usepackage{hyperref}
\def\BibTeX{{\rm B\kern-.05em{\sc i\kern-.025em b}\kern-.08em
    T\kern-.1667em\lower.7ex\hbox{E}\kern-.125emX}}
\graphicspath{{Journal/figures/evaluation}}
\newlength\Fcolumnseprule
\newcommand{\kstrongest}{$k$-strongest}

\usepackage{array}
\usepackage{subfig}
\usepackage{makecell}

\usepackage[printonlyused]{acronym}
\usepackage{enumitem}
\usepackage{wrapfig}
\usepackage{amsmath,amsfonts}
\usepackage{bbm} 
\usepackage{bm} 
\usepackage{fixmath} 
\definecolor{ForestGreen}{RGB}{34,139,34}

\usepackage{textcomp}
\usepackage{url}
\usepackage{verbatim}
\usepackage{graphicx}
\def\BibTeX{{\rm B\kern-.05em{\sc i\kern-.025em b}\kern-.08em
    T\kern-.1667em\lower.7ex\hbox{E}\kern-.125emX}}
    \usepackage{cleveref}
\usepackage{balance}
\usepackage{color, colortbl}
\usepackage{float}
\usepackage{accents}

\newcommand{\mycc}{\cellcolor{Gray}}
\acrodef{ICP}		[ICP]			{Iterative Closest Point}
\acrodef{CFAR}		[CFAR]			{Constant False-Alarm Rate}
\acrodef{FMCW}		[FMCW]			{Frequency-Modulated Continuous Wave}
\acrodef{TITLE}		[TBV-Radar SLAM]			{Trust But Verify loop candidates for robust radar SLAM}
\newcommand{\tbv}{TBV Radar SLAM}
\newcommand{\cfear}{CFEAR Radar Odometry}
\newcommand{\tbvshort}{TBV}

\newcommand{\changed}[1]{{#1}}

\author{Daniel Adolfsson, Mattias Karlsson, Vladimír Kubelka, Martin Magnusson, Henrik Andreasson
\thanks{Manuscript received: January, 10, 2023; Revised March, 13, 2023; Accepted April, 13, 2023. 
This paper was recommended for publication by Editor Javier Civera upon evaluation of the Associate Editor and Reviewers' comments.
This work was supported by Sweden's Innovation Agency under grant number 2021-04714 (Radarize), and 2019-05878 (TAMMP).} 
\thanks{The authors are with the MRO lab of the AASS research centre at \"Orebro University, Sweden. First author: \texttt{Dla.Adolfsson@gmail.com}, co-authors \texttt{firstname.lastname@oru.se}
}
\thanks{Digital Object Identifier (DOI): see top of this page.}
}
\title{\LARGE \bf 
TBV Radar SLAM -- trust but verify loop candidates
}

\begin{document}

%

\maketitle

  
\markboth{IEEE Robotics and Automation Letters. Preprint Version. Accepted April 2023}
{Adolfsson \MakeLowercase{\textit{et al.}}: TBV Radar SLAM -- trust but verify loop candidates} 

\maketitle

\begin{abstract}
Robust SLAM in large-scale environments requires fault resilience and awareness at multiple stages, from sensing and odometry estimation to loop closure. In this work, we present TBV (Trust But Verify) Radar SLAM, a method for radar SLAM that introspectively verifies loop closure candidates. \tbv{} achieves a high correct-loop-retrieval rate by combining multiple place-recognition techniques: tightly coupled place similarity and odometry uncertainty search, creating loop descriptors from origin-shifted scans, and delaying loop selection until after verification.
Robustness to false constraints is achieved by carefully verifying and selecting the most likely ones from multiple loop constraints. Importantly, the verification and selection are carried out after registration when additional sources of loop evidence can easily be computed.
We integrate our loop retrieval and verification method with a robust odometry pipeline within a pose graph framework. By evaluation on public benchmarks we found that \tbv{} achieves 65\% lower error than the previous state of the art. We also show that it generalizes across environments without needing to change any parameters. We provide the open-source implementation at \url{https://github.com/dan11003/tbv_slam_public}
\end{abstract}


\section{Introduction}
\IEEEPARstart{R}{obust} localization is key to enabling safe and reliable autonomous systems. Achieving robustness requires careful design at multiple stages of a localization pipeline, from environment-tolerant sensing and pose estimation, to place recognition and pose refinement. 
At each stage, a localization and mapping pipeline should be designed for fault awareness to detect failures and fault resilience to mitigate failures as they inevitably occur.

Today, active exteroceptive sensors such as lidar and radar are suitable when robust uninterrupted localization is required. Of these two, radar perception is significantly less affected when operating within dusty environments or under harsh weather conditions. It is currently debated how the sensing properties affect localization and mapping performance~\cite{hongSLAM_IJRR,replace_radar_lidar}. 
Compared to lidar, limited work focuses on robust and accurate radar SLAM, and none of the existing methods include introspective fault detection.


In this letter, we propose TBV (Trust But Verify) Radar SLAM -- a 2D pose-graph localization and mapping pipeline which integrates fault resilience and fault-aware robustness at multiple stages.
A radar-only odometry front-end adds pose nodes to the graph. In parallel, a robust loop-detection module adds loop closure constraints such that the SLAM back-end can optimize the graph to correct drift.
\tbv{} uses radar odometry (further only \emph{odometry}), place descriptors, and scan alignment measures to retrieve, verify, and select between loop constraints. We achieve a high correct-loop-retrieval rate by combining: a tightly coupled place similarity and odometry uncertainty search, creating place descriptors computed from origin-shifted scans, and by delaying loop selection until after verification.
Robustness, with a high loop detection rate, is achieved by unifying the process of place recognition and verification. Rather than rejecting candidates early during place recognition, we register and verify multiple loop constraints in parallel. Our verification combines place similarity, odometry consistency, and an alignment quality assessment automatically learned from odometry and scans.
\begin{figure}
    \centering
    \includegraphics[trim={0cm 0cm 0cm 0cm },clip,width=0.8\hsize]{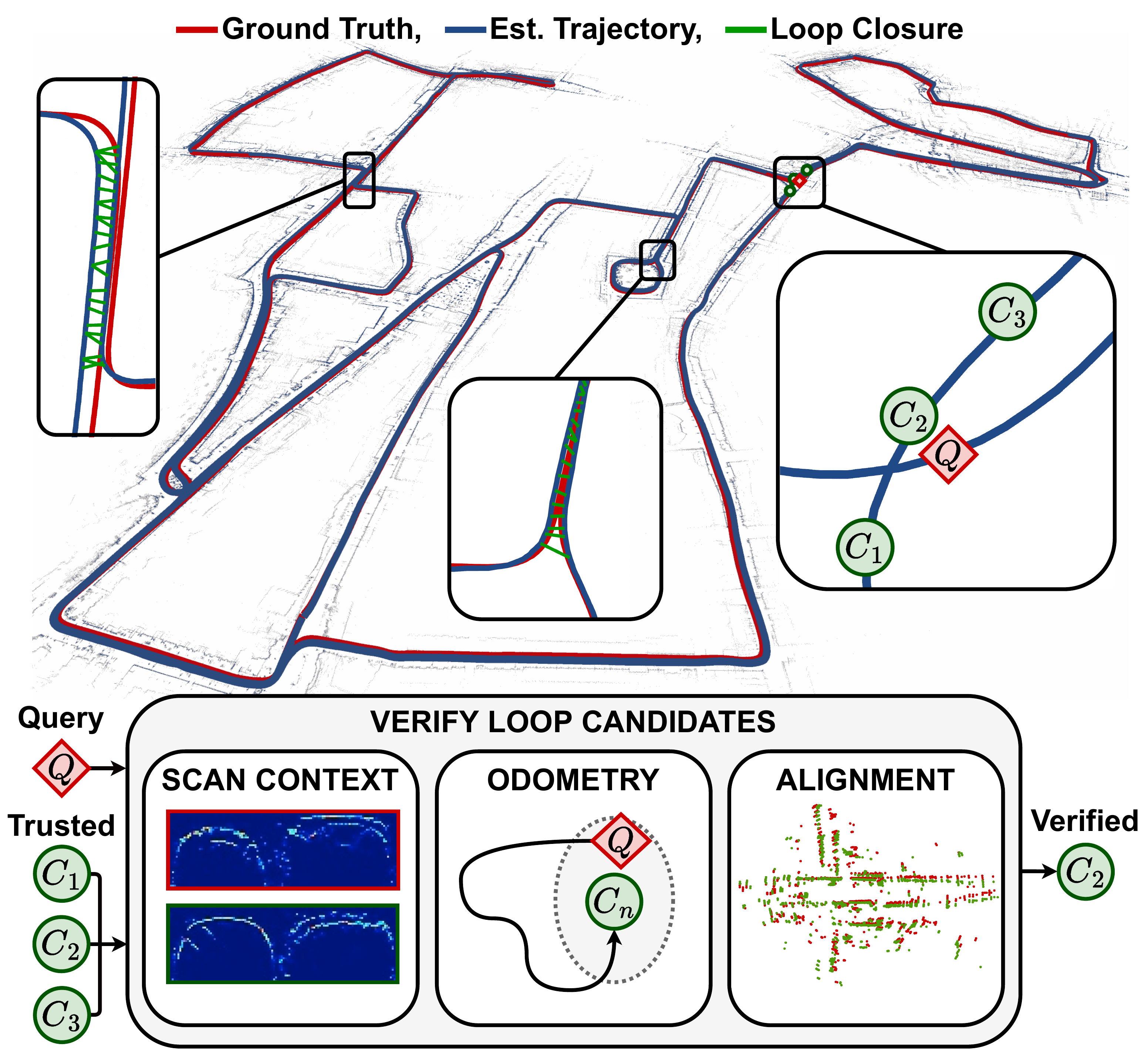}
    \caption{Overview and demonstration of \tbv{}} \url{https://tinyurl.com/TBVRadarSLAM}
    \label{fig:frontimage}
    \vspace{-0.4cm}
\end{figure}
We integrate our loop retrieval and verification module with a robust method for radar odometry, into a full-fledged SLAM pipeline visualized in Fig.~\ref{fig:frontimage}.

The contributions of this letter are:\changed{
\begin{itemize}

\item A combination of techniques for a high rate of correct loop retrievals, including coupled place similarity and odometry uncertainty search, and creating place descriptors from origin-shifted scans.

\item A verification step that jointly considers place similarity, odometry uncertainty, and alignment quality computed after registration. Verification and loop retrieval mutually benefit each other by selecting the loop constraint with the highest verified confidence.

\end{itemize}
We integrate these techniques with a robust odometry estimator into an online SLAM framework that pushes the state of the art in radar SLAM accuracy while generalizing between environments without retuning parameters.}
\begin{figure*}
\vspace{0.5cm}
    \centering
    \includegraphics[width=\hsize]{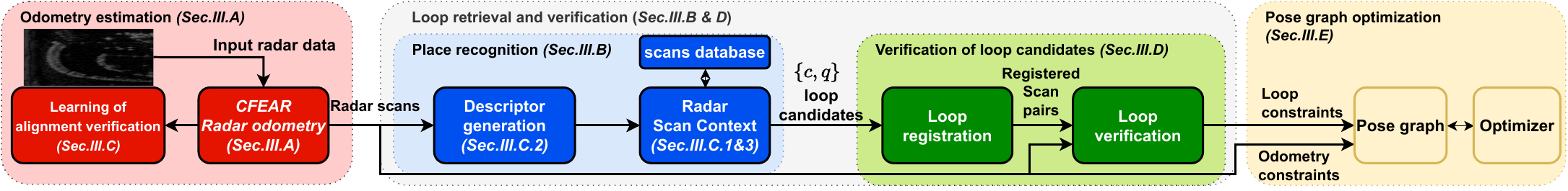}
    \caption{Overview of \tbv{}. The main contribution is the \changed{loop} retrieval and verification module. }
    \label{fig:tbv_overview}
    \vspace{-0.7cm}
\end{figure*}

\section{Related work}
\changed{
Recent progress in the development of spinning 2D radar has inspired work on odometry
estimation~\cite{cen2018,cen2019,barnes_masking_2020,barnes_under_2020,PhaRaO,kung2021normal,burnett2021radar,hongSLAM_IJRR,cfear_journal,what_goes_around,fmbm}, place recognition~\cite{saftescu_kidnapped_2020,kradarplusplus,gadd_unsupervised,gadd2020look,yin2021radar,RaLL},
topometric~\cite{kradarplusplus,replace_radar_lidar} and overhead imagery~\cite{self_sup_overhead} localization, and SLAM~\cite{hong2020radarslam,hongSLAM_IJRR,MAROAM}. 

Essential to both SLAM and topometric localization, is recognition of previously seen places, followed by careful refinement of the relative pose.
Hong et al.~\cite{hongSLAM_IJRR} use adaptive thresholding to filter data for computing place descriptors. Loop candidates are retrieved using M2DP~\cite{he2016m2dp}, and constraints are computed by registering sets of KLT tracker keypoints. Our work, however, brings a larger focus on the introspective verification of loop constraints. Martini et al.~\cite{kradarplusplus} proposed a teach-and-repeat framework with loop retrieval using ~\cite{saftescu_kidnapped_2020}.
%
Registration is carried out globally--without initial guess--by finding and minimizing the distance between the optimal set of landmark matches as described by Cen et al~\cite{cen2018}. In contrast, we use a fast local registration method, with an initial guess from place recognition as an initial alignment guess. However, we carefully verify registration success before accepting new loop constraints.

Verification of pose estimates is essential for safety and could have prevented an accident that occurred during an automated driving test~\cite{naoki_detection_of_loc}. 
Today, visual, lidar, and radar SLAM frequently adopts \textit{geometrical consistency} for loop verification. For visual SLAM, RANSAC is commonly used for verification~\cite{pastPresentFuture}. Thresholding based on the ICP quality measure (i.e., the squared point-to-point distance) is more frequently used for lidar~\cite{lio_sam,Vlaminck_2019_verify_icp}, and radar~\cite{Holder_2019}. In these cases, verification step runs in cascade, independently of accepting loop retrievals. In contrast, we present a unified approach that verifies loop retrieval and computed alignment jointly.
One family of methods proposes improved quality measures~\cite{ADOLFSSON2022104136,learning_to_detect,yin_a_failure_detection} for verification. We build upon~\cite{ADOLFSSON2022104136}. 
In contrast to the formerly mentioned methods, some works additionally consider the relation between measurements, for lidar map localization~\cite{naoki_detection_of_loc}, radar odometry~\cite{8917111}, and topometric localization~\cite{kradarplusplus}. 
Martini et al.~\cite{kradarplusplus} and Aldera et al.~\cite{8917111} use the landmark matching compatibility matrix~\cite{cen2018} to assess quality. Martini~\cite{kradarplusplus} reject place candidates based on the quality score of the matrix. Aldera~\cite{8917111} detects pose estimate failures from eigenvectors features of the compatibility matrix. Detection is learned with supervision from an external ground truth system. Our method use automatic generation of training data without additional sensors.

A complementary verification technique is \textit{temporal consistency}, which requires consistency over consecutive frames.
The technique has been adopted in lidar~\cite{behley-2018-suma} and visual SLAM~\cite{orb_slam_3}. We focus on verifying individual frames 
which enables loop detection from short overlapping segments.

Gomez-Ojeda et al.~\cite{Gomez_Ojeda_2019_loop_Stereo} use prior \textit{assumption of localization quality}, with a maximum displacement as part of loop verification. 
We do not embed hard constraints but penalize candidates that are unlikely given reasonable assumption of drift.
Finally, some methods use an early \textit{risk assessment} 
by measuring how range measurements could constrain registration. 
A low level of constraints in one direction suggests registration was~\cite{Gomez_Ojeda_2019_loop_Stereo}, or could be unstable~\cite{8462890}. 
The measure has been used to
abstain from~\cite{hongSLAM_IJRR} or prioritize inclusion of loop closures~\cite{denniston_darpa_loop_closure_prior}, or plan motion accordingly~\cite{castellano-2022-alignability}. We instead rely on geometric verification after registration.

\label{sec:rw-radar} 
}


\section{\tbv{}}

An overview of \tbv{} is presented in Fig.~\ref{fig:tbv_overview}. In this section, we detail the components including \cfear{} (Sec.~\ref{sec:cfear_radarodometry}), place recognition (Sec.~\ref{sec:place_recognition}), verification of loop candidates (Sec.~\ref{sec:loop}),  and pose graph optimization (Sec.~\ref{sec:pose_graph_optimization}). The self-supervised training of alignment verification (Sec.~\ref{sec:self_supervised_alignment_verification}) runs once during a learning phase and is not required for online SLAM.

\subsection{\cfear{}}
\label{sec:cfear_radarodometry}
We use the radar odometry pipeline \textit{\cfear{}}~\cite{cfear_journal} (specifically the \textit{CFEAR-3} configuration). This method takes raw polar radar sweeps as input and estimates sensor pose and velocity. 
As an intermediate step, the method computes a set of radar peaks $P^t$ and a set of oriented surface points $\mathcal{M}^t$ at time $t$. \changed{These sets are reused for other modules.}
CFEAR filters the radar data by keeping the \kstrongest{} range bins per azimuth.
The filtered point set is compensated for motion distortion and transformed into Cartesian space, - this is also the point set from which we extract radar peaks ($P^t$). A grid method is then used to compute a set of oriented surface points $\mathcal{M}^t$ from the filtered set. Odometry is estimated by finding the pose $\mathbf{x}^t$ in SE(2) which minimizes the sum of surface point distances between the current scan $\mathcal{M}^t$ and the $|\mathcal{K}|$ latest keyframes $\mathcal{M}^\mathcal{K}$ jointly as
\begin{equation}
\label{eq:scan_to_multikeyframes}
 f(\mathbf{\mathcal{M}}^{\mathcal{K}},\mathbf{\mathcal{M}}^t,\mathbf{x}^t) =\sum_{\hspace{-0.3cm}k\in\mathcal{K}}\hspace{-0.5cm}\sum_{\hspace{0.5cm}\forall \{i,j\}\in \mathcal{C}_{corr}} \!\!\!\!\!\!\!\!\!w_{i,j}\rho\big(g(m^k_j, m^t_i,\mathbf{x}^t)\big),
\end{equation}
where $w_{i,j}$ are surface point similarity weights, $\rho$ is the 
Huber loss function, and $g$ is the 
pairwise distance between surface points ($m$) in the correspondence set $\mathcal{C}_{corr}$.  
A keyframe $k\in\mathcal{K}$ is created when the distance to the previous exceeds $1.5$~m. This technique reduces drift and removes excessive scans acquired when the robot remains stationary. Upon creation of a new keyframe, odometry constraints are created from the relative alignment between the current pose and the latest keyframe. Odometry constraints are added to the set $\mathcal{C}_{odom}$ and used to correct trajectory (via pose graph optimization) as detailed in Sec.~\ref{sec:pose_graph_optimization}.

\subsection{Place recognition}
\label{sec:place_recognition}
We attempt to unify the stages of loop closure, from detection to constraint verification.
Accordingly, we retrieve, register, and verify multiple candidates, of which one \changed{can be selected}.
For that reason, we \changed{don't} discard potential loop candidates based on place similarity. Instead, we trust multiple candidates to be meaningful until verified. At this point, we select the verifiably best \changed{candidate}, if such \changed{exists.}

\subsubsection{Scan Context}
We build upon the place recognition method \textit{Scan Context} by Kim et al.~\cite{ gkim-2018-iros,gskim-2021-tro}. Their method detects loops and relative orientation by matching \changed{the most recently acquired scan (referred to as query $q$)} with candidates ($\{c\}$) stored in a database using scan descriptors. Scenes are encoded by their polar representations into 2D descriptors 
$\mathbf{I}_{ring \times sec}$. The 2D descriptor is in turn used to create a rotation-invariant 1D descriptor (ring key) via a ring encoding function. Loops are detected via a two-step search: First, the query 1D ring key is  matched with the top 10 candidates via a fast nearest neighbor (NN) search. Second, for each of the top candidates, a sparse optimization finds the relative rotation that minimizes a distance metric $d_{sc}(\mathbf{I}^q,\mathbf{I}^{c})$: the sum of cosine distances between descriptors columns. The candidate $c$ with the lowest score ($d_{sc}$) which does not exceed the fixed threshold $\tau$ is selected as loop candidate.
\begin{equation}
\label{eq:sc_sim}
\begin{split}
    c =  \underset{c \in \mathcal{C}_{and}}{\operatorname{arg min}} \: d_{sc}(\mathbf{I}^q,\mathbf{I}^{c}), s.t.\: d_{sc} < \tau. 
\end{split}
\end{equation}
\changed{In our implementation, query descriptors are created, matched, and stored in the database for each keyframe.
}

\subsubsection{Descriptor generation}
\label{sec:descriptor_generation}
As described in~\cite{gskim_2020_mulran,MAROAM}, a raw polar representation such as those produced by a spinning 2D radar, can be used directly as a Scan Context descriptor. However, we believe that doing so poses multiple challenges, including sensor noise, motion distortion, scene dynamics, and translation sensitivity. Thus, we 
create our descriptor from multiple filtered and motion-compensated scans.
Conveniently, such processed scans are already provided by the CFEAR.
We aggregate the peak detections from keyframe $q$ and its two neighbors in the odometry frame. 


Having the radar scan represented as a sparse point cloud in Cartesian space allows us to address translation sensitivity in place recognition by applying the data augmentation step (Augmented PC) from~\cite{gskim-2021-tro} before computing place descriptors. 
We perform data augmentation by shifting the sensor origin, i.e. by transforming $\mathcal{P}^{q}$ with $\pm$2 and $\pm$4~m lateral translation offsets. The original, and the 4 augmented point clouds, are each used to compute and match descriptors, after which the best matching pair of query/candidate is selected. Note that by using the aggregated sparse point cloud, rather than the dense raw radar scan, we can efficiently compute all augmentations and corresponding descriptors. As such, the main computational load from the augmentation technique is due to matching of additional descriptors and not the computation of these.
The descriptor itself is created by populating the Scan Context $\mathbf{I}$ with radar intensity readings. 
Specifically, for each grid cell $\mathbf{I}(i,j)$ we sum the intensity of all coinciding point intensities (of radar peaks) divided by $1000$. Empty cells are set to $\mathbf{I}(i,j)=-1$, which we found increased descriptiveness compared to $\mathbf{I}(i,j)=0$ \changed{by distinguishing unoccupied from possibly occupied cells.}

\subsubsection{Coupled odometry/appearance matching}
\label{sec:coupled_matching}
\changed{
When retrieving loop candidates, odometry uncertainty information can be used to directly reject unlikely loop constraints.
If the likelihood of the loop constraint $\mathbf{x}^{q,c}_{loop}$ (given the estimated odometry trajectory $\mathbf{v}_{odom}^{c:q}$ between $c$ and $q$ ) is close to zero:
\begin{equation}
\label{eq:odom}
p(\mathbf{x}^{q,c}_{loop}\:|\mathbf{v}_{odom}^{c:q}) \approx 0,     
\end{equation}
the proposed constraint is false and can be discarded.} 
\changed{We choose however to jointly consider the odometry uncertainty and place similarity to obtain more robust loop-retrievals. 

Accordingly, we propose a coupled place similarity / odometry uncertainty search, which combines Eq.~\ref{eq:sc_sim} and Eq.~\ref{eq:odom}.} Candidates are thus selected jointly by the similarity of appearance $d_{sc}(\mathbf{I}^q,\mathbf{I}^{c})$ and the similarity of odometry $d^{q,c}_{odom}$:
\begin{equation}
\label{eq:coupled_sim}
\begin{split}
    c =  \underset{c \in \mathcal{C}}{\operatorname{arg min}} \:d_{sc}(\mathbf{I}^q,\mathbf{I}^{c}) + d^{q,c}_{odom},\\
    d^{q,c}_{odom} = 1 - p(\mathbf{x}^{q,c}_{loop}\:|\mathbf{v}_{odom}^{c:q})\:.
\end{split}
\end{equation}
\changed{
We choose to estimate the likelihood in Eq.~\ref{eq:odom} using a Gaussian distribution $p(\mathbf{x}^{q,c}_{loop}\:|\mathbf{v}_{odom}^{c:q})=\exp{(-\frac{t_{err}^2}{2\sigma^2})}$.}
\begin{equation}
t_{err} = \frac{\max(||transl(\mathbf{x}^q)-transl(\mathbf{x}^c)||- \epsilon, 0)}{ dist(\mathbf{v}_{odom}^{c:q})}.
\end{equation}
Here, $transl$ is the translation component, $\epsilon$ is the expected maximum spacing between loop candidates (fixed to $\epsilon=5$ i.e. slightly higher than the lateral augmentation distance), and $dist(\mathbf{v}_{odom}^{c:q})$ is the traversed distance between the query and loop candidate estimated by the odometry. Note that $t_{err}$ quantifies the \textit{relative final position error}, thus $\sigma$ can be chosen according to expected odometry quality to penalize unlikely loops.
We refrained, however, from making strong assumptions on odometry quality, and fixed $\sigma=0.05$; i.e., a pessimistic assumption of $5\%$ relative translation error.

Note that the two-step search of Scan Context requires that odometry uncertainty is integrated already in the 1D NN search. We do this by extending all 1D descriptors (of size $ring=40$) with odometry similarity scores ($d_{odom}$) as an extra element. 
 ($d_{odom}$) is scaled with a factor $(ring/4)$ to balance odometry uncertainty and appearance similarity.


\subsection{Automatic learning of alignment verification}
\label{sec:self_supervised_alignment_verification}
To improve loop closure verification, we build upon the system \textit{CorAl}~\cite{ADOLFSSON2022104136} which learns to detect alignment errors between two registered scans. CorAl allows us to determine if a loop constraint is correct by formulating loop constraint verification as a misalignment detection problem.
Specifically, a loop (between scan nr $q$ and $c$) is verified as \textit{correct} only if the scans 
are correctly aligned via the relative alignment $\mathbf{x}^{q,c}_{loop}$.
During the learning phase, CorAl automatically generates labeled training data. 
The input to CorAl is pairs of odometry estimates, radar peak detections ($\mathcal{P}$), and computed sets of oriented surface points ($\mathcal{M}$). These entities are used to extract alignment quality residuals from which alignment quality can be assessed.
After the learning phase, CorAl can verify loops by detecting alignment errors (caused e.g. by 
heavy motion distortion or incorrect convergence). CorAl also aids in distinguishing between places that differ by small geometric details.
We generate training data by repeating the following process for each pair of consecutive keyframes.
Positive training labels ($y_{aligned}\!=\!true$) and training data $X_{quality}$ are computed using the scan alignment 
provided by the odometry. For each pair, the alignment quality measures in Eq.~\ref{eq:quality_measure} are extracted. 
Negative training labels ($y_{aligned}\!=\!false$) and training data are extracted similarly. However, before extracting the alignment quality, an error is induced in the alignment in both translation and rotation. This allows us to learn the detection of different types of errors. Specifically, we distribute 12 translation errors symmetrically in either the positive or negative x or y-axis. We use 4 small ($\pm 0.5$~m), 4 medium ($\pm 1$~m) and 4 large ($\pm 2$~m) errors. To these errors, we additionally induce a clockwise rotation with matching rotation errors: small ($0.5^\circ)$, medium ($ 2^\circ$) or large ($15^\circ$). 
Note that the class ratio 1:12, between positive to negative training is alleviated during learning by assigning weights according to the inverse of class frequency. 
\subsubsection{Alignment measure}
\label{sec:alignmeasure}
We extract the following alignment measures between each pair of scans:
\begin{equation}
\label{eq:quality_measure}
\setlength\abovedisplayskip{8pt}
    \mathbf{X}_{quality} = [H_{j} \: H_{s} \: H_{o} \: C_f \: C_o \: C_{a} ~1]^T.
    \setlength\belowdisplayskip{8pt}
\end{equation}
The joint entropy ($H_j$) and separate entropy ($H_s$) are average per-point differential entropies, extracted from point cloud pairs of radar peak detections ($\mathcal{P}^q,\mathcal{P}^{c}$). \changed{Average per-point entropy is a measure of point cloud \textit{crispness} -- lower crispness following registration indicates misalignment. }
These metrics are described in-depth in~\cite{ADOLFSSON2022104136}. 
We complement these measures with a measure of overlap $H_o$: ($H_{overlap}$), defined as the portion of peaks in $\mathcal{P}^q$ or $\mathcal{P}^{c}$ with a neighboring point within the radius $r$ in the other point cloud.

In this work, we combine these CorAl measures with additional ones, extracted from ($\mathcal{M}^q,\mathcal{M}^{c}$), i.e. from pairs of scans represented as oriented surface points. The measures are obtained from the registration cost(Eq.~\ref{eq:scan_to_multikeyframes}), but with a single keyframe and with the point-to-line cost function ($g_{P2L}$~\cite{cfear_journal}).
Note that these measures are already computed during the final iteration of the registration, and this step brings little computational overhead. Specifically, from Eq.~\ref{eq:scan_to_multikeyframes} we reuse $C_f$: $f(\mathcal{M}^q, \mathcal{M}^{c},\mathbf{x}^{q,c}$), the number of correspondences (overlap) $C_o$: $|\mathcal{C}_{corr}|$, and average surface point set size $C_{a}$: $1/2(|\mathcal{M}^q| + |\mathcal{M}^{c}|)$. The intuition of combining these quality measures is that the CorAl measures (which use small-region data association) are specialized in detecting small errors whereas the CFEAR measures are more suitable for larger alignment errors.
\changed{We refer to~\cite{ADOLFSSON2022104136} for details.}

\subsubsection{Assessing alignment}
\label{sec:assess}
Once training data has been computed, we train a logistic regression classifier
\begin{equation}
\setlength\abovedisplayskip{5pt}
 p_{align} =  1/(1+e^{-d_{align}}),\quad
 d_{align} = \boldsymbol{\beta} \mathbf{X}_{quality},\label{eqn:d_align}
 \setlength\belowdisplayskip{5pt}
\end{equation}
where $\boldsymbol{\beta}_{1\times 7}$ are the learned model parameters. We train on discrete alignment classification here as we consider all visible errors to be undesired. However, $d_{align}$ is passed to our loop verification module rather than $p_{align}$. We found $d_{align}$ to be more suitable, as the sigmoid output $p_{align}$ is insensitive to alignment change close to $0$ or $1$.

\subsection{Verification of loop candidates}
\label{sec:loop}
We allow for multiple competing loop candidates $c_k$ per query $q$ as illustrated in Fig.~\ref{fig:frontimage}. Each of the $N_{cand}=3$ best matching pairs $\{(q,c_k)\}$ provided by the place recognition module is used to compute and verify potential loop constraints. A constraint is computed by finding the relative alignment $\mathbf{x}^{q,c_k}_{loop}$ that minimizes Eq.~\ref{eq:scan_to_multikeyframes} i.e. the distance between correspondences, similar to the odometry module. As an initial guess, we use the relative orientation provided by the place recognition module. If the loop candidate was retrieved from a match with an augmented query scan, the corresponding augmented lateral offset is used together with the rotation as an initial guess. Note that the local registration method is motivated by the access to an initial guess, required for convergence. After registration, we extract and assess the alignment quality $d_{align}=\boldsymbol{\beta} \mathbf{X}^{q,c_k}_{quality}$ following the procedure in Sec.~(\ref{sec:alignmeasure}\&\ref{sec:assess}). Each constraint is finally verified by combining the Scan Context distance $(d_{sc}$) with odometry uncertainty ($d_{odom}$) and alignment quality ($d_{align}$) with a logistic regression classifier:
\begin{equation}
\setlength\abovedisplayskip{5pt}
\begin{split}
    &y_{loop}^{q,c_k} = \frac{1}{1+e^{-\boldsymbol{\Theta}\mathbf{X}^{q,c_k}_{loop}}}, \:\:s.t.\:\: y_{loop}^{q,c_k}> y_{th},\\
    &\mathbf{X}^{q,c_k}_{loop} = [d_{odom} \: d_{sc} \: d_{align} \: 1]^T. 
    \end{split}
    \setlength\belowdisplayskip{5pt}
\end{equation}
The model parameters $\boldsymbol{\Theta}$ can be learned via ground truth loop labels, or simply tuned as the 4 parameters have intuitive meaning. $y_{th}$ is the sensitivity threshold -- we rarely observe false positives when fixed to 0.9.
\begin{figure*}[b!]
\vspace{-0.85cm}
	\centering
	\newcommand\figsize{0.1425\hsize}
	\newcommand\trimleft{1}
	\newcommand\trimright{0.4}
	\begin{center}
		\subfloat[T.1]{\includegraphics[trim={\trimleft cm 0cm  \trimright cm 0cm},clip,width=\figsize]{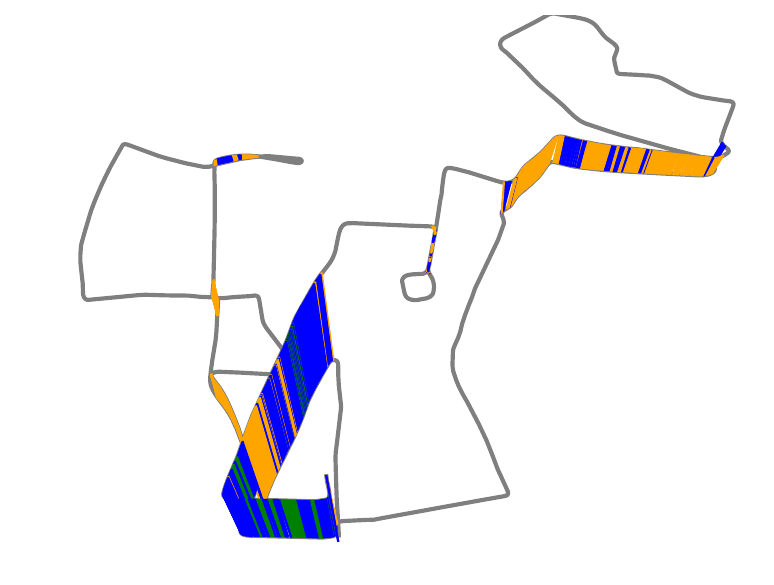}\label{fig:1_qualitative}}
		\subfloat[T.2]{\includegraphics[trim={\trimleft cm 0cm \trimright cm 0cm},clip,width=\figsize]{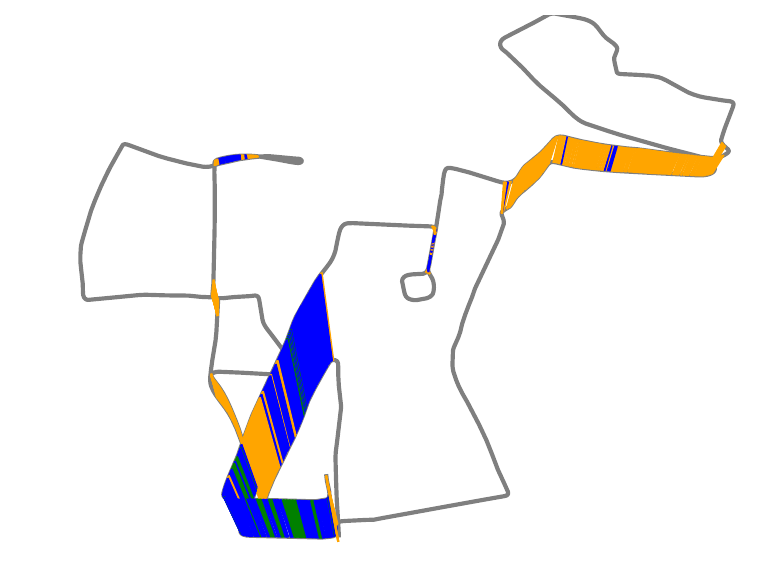}\label{fig:2_qualitative}}
		\subfloat[T.3]{\includegraphics[trim={\trimleft cm 0cm \trimright cm 0cm},clip,width=\figsize]{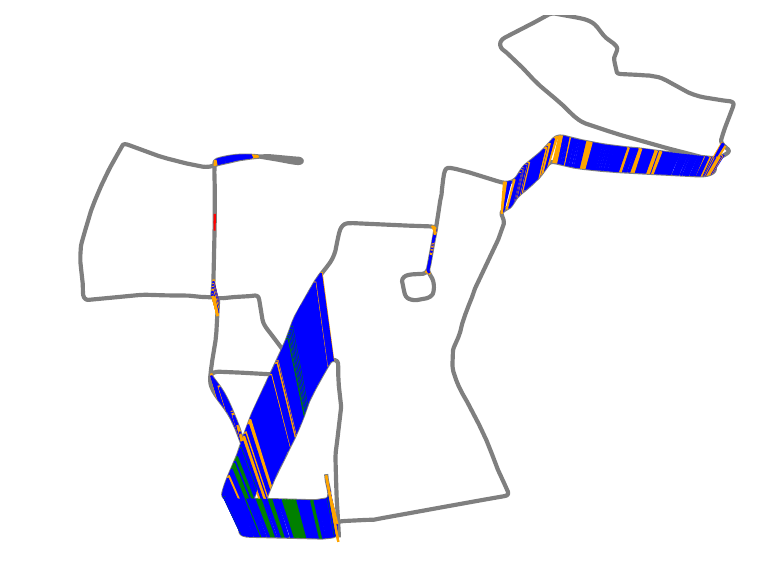}\label{fig:3_qualitative}}
		\subfloat[T.4]{\includegraphics[trim={\trimleft cm 0cm \trimright cm 0cm},clip,width=\figsize]{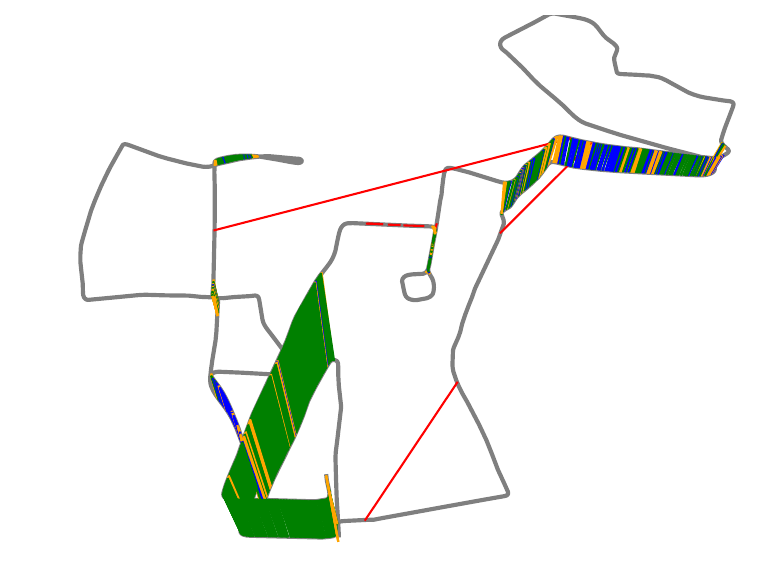}\label{fig:4_qualitative}}
		\subfloat[T.5]{\includegraphics[trim={\trimleft cm 0cm \trimright cm 0cm},clip,width=\figsize]{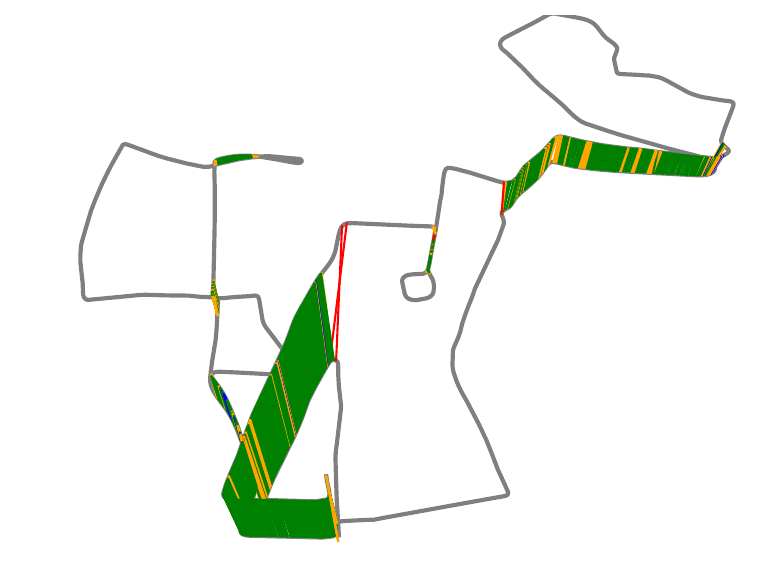}\label{fig:5_qualitative}}
		\subfloat[T.6]{\includegraphics[trim={\trimleft cm 0cm \trimright cm 0cm},clip,width=\figsize]{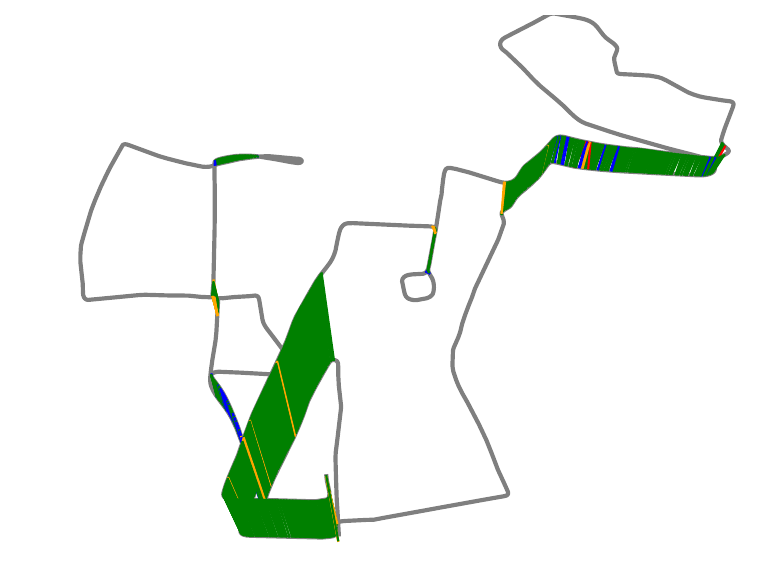}\label{fig:6_qualitative}}
		\subfloat[T.8]{\includegraphics[trim={\trimleft cm 0cm \trimright cm 0cm},clip,width=\figsize]{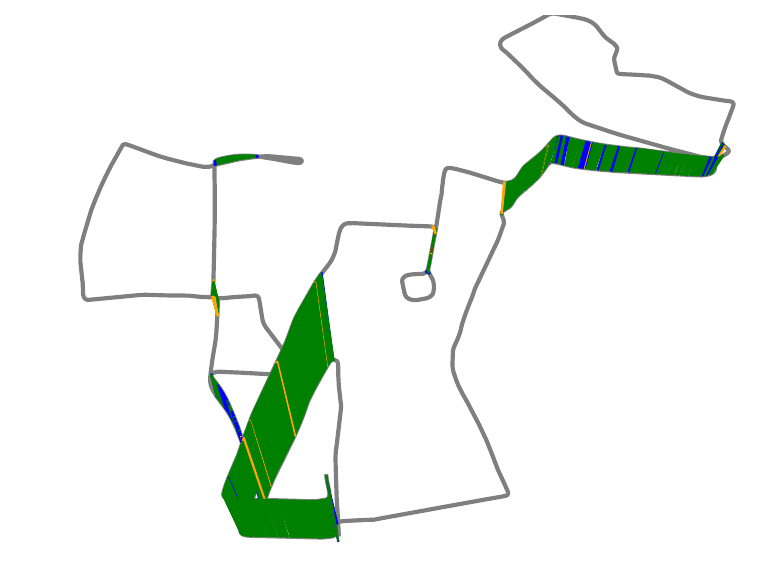}\label{fig:7_qualitative}}
		\vspace{-0.2cm}
		\caption{
  We visualize the impact of loop retrieval and verification for the methods in \textbf{Sec.~\ref{sec:ablation}} within the Oxford dataset.
  All potential loops are colored and ordered according to loop retrieval correctness and confidence ($y_{loop}^{q,c_k}$): red (dangerous failure -- false loop with undesired high confidence), orange (safe failure -- false loop but with desired low confidence), blue (safe failure, correct loop with undesired low confidence),  green (success, correct loop with desired high confidence).
		 \label{fig:qualitative_loops}}
	\end{center}
	\vspace{-0.0cm}
\end{figure*}

We investigated two strategies for selecting loop closures after a successful verification: (i) We select the first candidate retrieved from the place recognition module ($N_{cand}=1$) -- the lowest Scan Context distance score $d_{sc}$. (ii) We use $N_{cand}=3$ candidates and select the best candidate according to our verifier -- the highest probability $y_{loop}^{q,c_k}$. The intuition for strategy (i) is that the first retrieved place candidate is often the best guess, without considering registration. However, there are cases where one of the latter candidates is preferred. For example, registration of query and the first candidate may fail, or subtle scene differences that distinguish places can be hard to detect until a more thorough local analysis of alignment has been carried out. Thus, selecting the verifiably better loop constraint candidate is desired.
We compare these two strategies in Sec.~\ref{sec:ablation} (~\ref{para6} and ~\ref{para8})  Once a loop constraint has been verified, it is added to the set $\mathcal{C}_{loop}$.


\subsection{Pose Graph Optimization}
\label{sec:pose_graph_optimization}
We correct the odometry by solving a sparse least squares optimization problem. We do so by minimizing Eq.~\ref{eq:slam}, using the odometry and loop constraints $\mathcal{C}_{odom}$, $\mathcal{C}_{loop}$:\changed{
\begin{equation}
\setlength\abovedisplayskip{3pt}
\label{eq:slam}
\begin{split}
    J(\mathbf{Y}) = \!\!\!\!\!\!\! \underbrace{\sum_{ (i,j) \in \mathcal{C}_{odom}} \!\!\!\!\!\!\mathbf{e}_{i,j}^T\mathbf{C}^{-1}_{i,j}\mathbf{e}_{i,j}}_{\text{odometry constraints}} 
    +\!\!\!\! \underbrace{\sum_{ (i,j) \in \mathcal{C}_{loop}} 
    \!\!\!\!\!\!\alpha\rho( \mathbf{e}_{i,j}^T\mathbf{C}^{-1}_{i,j}\mathbf{e}_{i,j})}_{\text{loop constraints}}.
\end{split}
\setlength\belowdisplayskip{3pt}
\end{equation}
}$\mathbf{Y}=[\mathbf{y}_1\:\mathbf{y}_2...\mathbf{y}_n]$ is the vector of optimization parameters,
$\mathbf{e}\!=\!\mathbf{y}_{i,j\!}-\mathbf{x}_{i,j}$ is the difference between parameter and constraint, \changed{$\rho$ is the robust Cauchy loss function},
$\mathbf{C}$ is the covariance matrix. 
We used two strategies to compute covariance; fixed: computed from registration error statistics $\mathbf{C} = diag([v_{xx}=\text{1e-2}, v_{yy}=\text{1e-2}, v_{\theta\theta}=\text{1e-3}])$; dynamic: obtained from the Hessian, approximated from the Jacobian ($\mathbf{C}=(\mathbf{H})^{-1}\approx(\mathbf{J}^T\mathbf{J})^{-1}$) of the registration cost  (Eq.~\ref{eq:scan_to_multikeyframes}). Note that the dynamically obtained $\mathbf{C}$ was tuned by a factor $\gamma$ to provide realistic uncertainties, discussed in~\cite{barnes_masking_2020}. Loop constraints are optionally scaled by
\changed{($\alpha=\text{5e-5}$). }
This factor retains the high odometry quality through pose graph optimization and alleviates the need for a more accurate but computationally more expensive multi-scan loop registration.

Finally, we solve $\operatorname{arg min}_\mathbf{Y} J(\mathbf{Y})$ using 
Levenberg-Marquardt. Note that we do not need a robust back-end to mitigate outlier constraints (such as dynamic covariance scaling~\cite{dyn_cov_scaling} or switchable loop constraints~\cite{switchable_constraints}).

\section{Evaluation}
\label{sec:evaluation}

We evaluate our method on the \texttt{Oxford}~\cite{RadarRobotCarDatasetICRA2020} and \texttt{MulRan}~\cite{gskim_2020_mulran} datasets. Both datasets were collected by driving a car with a roof-mounted radar. 
The \texttt{Oxford} dataset contains 30 repetitions of a 10~km urban route. The \texttt{MulRan} dataset has a wider mix of routes; from structured urban to partly \changed{feature-poor} areas such as open fields and bridge crossings. \changed{In \texttt{MulRan}, places are frequently revisited multiple times with similar speeds in the same direction and road lane. This is not the case in the \texttt{Oxford} dataset where loop closure is significantly harder, and the mapping scenario is more realistic.} From these datasets, we selected the previously most evaluated sequences, see Tab.~(\ref{tab:TabOxfordATE}\&\ref{tab:TabMulRanATE}).
The radars used are Navtech CTS350-X, configured with 4.38~cm resolution in the \texttt{Oxford} dataset, and CIR204-H, with 5.9~cm resolution in the \texttt{MulRan} dataset. We use standard parameters for CFEAR-3~\cite{cfear_journal}, CorAl~\cite{ADOLFSSON2022104136}, and Scan Context ~\cite{gkim-2018-iros}, except where explicitly mentioned. 

\subsection{Run-time performance}
\changed{After the initial generation of training data, the full pipeline including odometry and loop closure runs with online capability. 
Pose graph optimization runs in a separate thread, either continuously as new verified loop constraints are computed, or once at the end. Run-time performance (measured on a 2015 mid-range consumer desktop CPU: i7-6700K) is as follows:
Odometry: 37\,ms/scan; Generation of training data: 236\,ms/keyframe pair; Pose graph optimization (once in the end with only odometry as prior): 992\,ms @ 4k keyframes; Loop closure (128\,ms @ $N_{cand}=3$); Descriptor generation: 1~ms/keyframe; Detect loops: 25\,ms/keyframe; Registration: 6\,ms/candidate; Verification 19\,ms/candidate.}

\afterpage{
\setcounter{figure}{3}
\begin{figure}[t]
	\vspace{0cm}
	\centering
	\newcommand\figsize{0.49\hsize}
	\begin{center}
		\subfloat[][\texttt{Oxford}]{\includegraphics[trim={0.2cm 0cm 0.5cm 0.65cm},clip,width=.49\hsize]{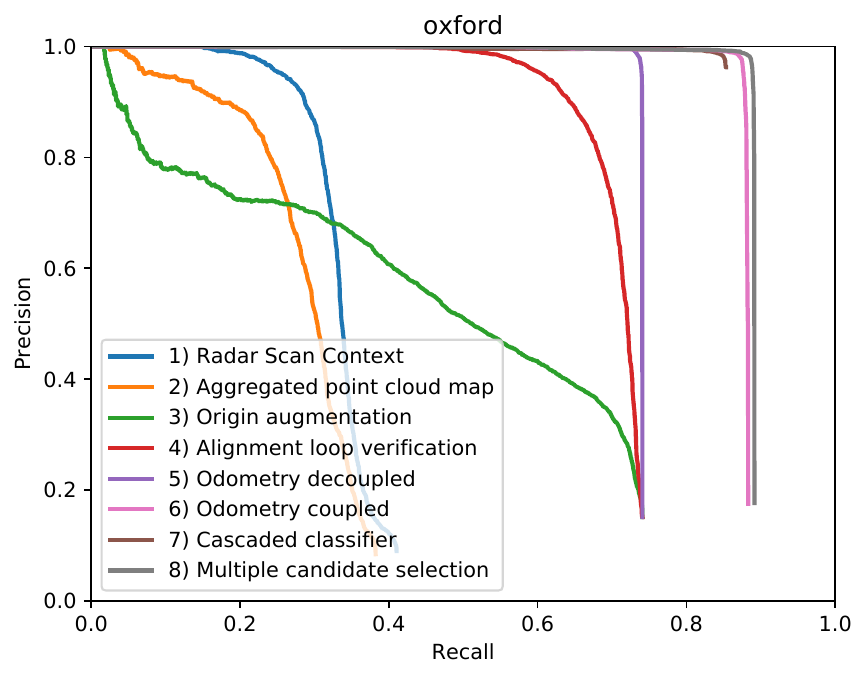}\label{fig:oxford_loop_closure_ablation}}
		\subfloat[][\texttt{MulRan}]{\includegraphics[trim={0.0cm 0cm 0.5cm 0.65cm},clip,width=0.49\hsize]{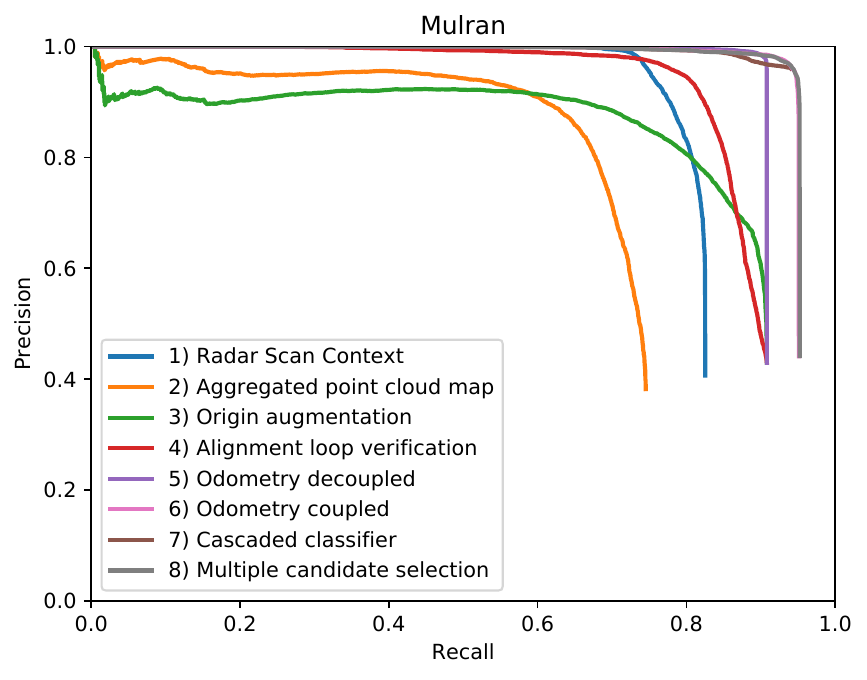}\label{fig:mulran_loop_closure_ablation}}
  \vspace{-0.2cm}
		\caption{Loop closure performance over all sequences. \label{fig:loop_closure_ablation}}
	\end{center}
	\vspace{-0.6cm}
\end{figure}}

\subsection{Ablation study -- loop detection}
\label{sec:ablation}
In this section, we evaluate the effect of the various aspects of our loop detection pipeline.
Loops are classified as correct if the difference between estimated alignment and ground truth does not exceed $4$~m or $2.5^\circ$. This error limit was set slightly higher than the largest pose error that we found in the ground truth at loop locations. 
While we would like to demonstrate the full capability of our system by analysis of smaller errors, we found ground truth accuracy to be a limiting factor. Note that the ground truth quality has previously been discussed as a limitation~\cite{kung2021normal}.

We compare the impact of each technique summarized below. Note that a later technique in the list either adapts or replaces the former technique. 

\definecolor{Gray}{gray}{0.9}
\begin{table*}[b!]
\centering
  \begin{adjustbox}{width=\textwidth}
\begin{tabular}{p{3ex}l|l|l|p{13mm}lllllll|ll}
\cline{2-13}
& \textbf{Type/modality} & \textbf{Method} & \textbf{Eval.} &  \texttt{10-12-32}\newline Training & \texttt{16-13-09} & \texttt{17-13-26} & \texttt{18-14-14} & \texttt{18-15-20} & \texttt{10-11-46} & \texttt{16-11-53} & \texttt{18-14-46} & Mean\\
\cline{2-13}
\ldelim\{{6.5}{30pt}[\rotatebox{90}{\hspace{-2mm}ATE}]&&&&&&&&&&&&&  
\\[-2ex]
& SLAM/camera & ORB-SLAM2~\cite{7946260} &~\cite{hongSLAM_ARXIV } & 7.96 & 7.59 & 7.61 &  24.63 & 12.17 & 7.30      & 3.54       &  9.72      &  10.07 \\
\cline{2-13}\\[-2ex]
& Odometry/Radar & CFEAR-3 (odometry used) & & 7.29 &      23.32 &      15.58 &      20.95 &      20.02 &      16.87 &      15.47 &      28.58 &  18.51\\
\cline{2-13} \\[-2ex]
&  \mycc \textbf{SLAM/Radar} &  \mycc RadarSLAM-Full~\cite{hongSLAM_IJRR} &\mycc ~\cite{hongSLAM_ARXIV} &\mycc 9.59        &\mycc 11.18       &\mycc 5.84       &\mycc 21.21        &\mycc 7.74       &\mycc 13.78        &\mycc 7.14        &\mycc 6.01       &\mycc 10.31\\
&  \mycc \textbf{SLAM/Radar} &  \mycc MAROAM~\cite{MAROAM} &\mycc {Author} &\mycc 13.76        &\mycc 6.95       &\mycc 8.36       &\mycc 10.34        &\mycc 10.96       &\mycc 12.42        &\mycc 12.51        &\mycc  7.71       &\mycc 10.38\\


&\mycc  \textbf{SLAM/Radar} &\mycc \tbv{}-dyn-cov-T.8 (ours)  &\mycc &\mycc \underbar{4.22} &\mycc \underbar{4.30}  &\mycc \textbf{3.37} &\mycc \underbar{4.04} &\mycc \underbar{4.27} &\mycc \underbar{3.38} &\mycc \underbar{4.98} &\mycc \textbf{4.27} &\mycc \underbar{4.10}  \\
&\mycc  \textbf{SLAM/Radar} &\mycc \tbv{}-T.8 (ours)  &\mycc &\mycc \textbf{4.07} &\mycc \textbf{4.04} &\mycc \underbar{3.58} &\mycc \textbf{3.79} &\mycc \textbf{3.83} &\mycc       \textbf{3.14} &\mycc \textbf{4.39} &\mycc \underbar{4.33} &\mycc \textbf{3.90} \\
\cline{2-13}\\[-2ex]
\cline{2-13} 
\ldelim\{{6.5}{30pt}[\rotatebox{90}{\hspace{-2mm}Drift}]&&&&&&&&&&&&&  
\\[-2ex]
& SLAM/Lidar & SuMa (Lidar - SLAM)~\cite{behley2018rss}  & ~\cite{hongSLAM_ARXIV} & $1.1/0.3^p$ & $1.2$/$0.4^p$ & $1.1/0.3^p$ & $0.9/0.1^p$ & $1.0/0.2^p$  & $1.1/0.3^p$        & $0.9/0.3^p$        & $1.0/0.1^p$        &  $1.03/0.3^p$\\
\cline{2-13} \\[-2ex]
&   \textbf{Odometry/Radar} & CFEAR-3-S50~\cite{cfear_journal} (Offline) &  ~\cite{cfear_journal} & \textbf{1.05/0.34} & \textbf{1.08/0.34} & \underbar{1.07/0.36} & \textbf{1.11/0.37} & \textbf{1.03}/\underbar{0.37} & \textbf{1.05}/\underbar{0.36} & \textbf{1.18/0.36} & \underbar{1.11}/\textbf{0.36} &  \textbf{1.09}/\textbf{0.36}\\

& Odometry/Radar & CFEAR-3 (odometry used) & & 1.20/0.36  & 1.24/0.40  & 1.23/0.39  & 1.35/0.42  & 1.24/0.41  & 1.22/0.39  & 1.39/0.40  & 1.39/0.44  & 1.28/0.40\\
\cline{2-13}
\\[-2ex]
 &\mycc  \textbf{SLAM/Radar} &\mycc RadarSLAM-Full~\cite{hongSLAM_IJRR} &\mycc ~\cite{hongSLAM_ARXIV} &\mycc $1.98/0.6$        &\mycc $1.48/0.5$        &\mycc $1.71/0.5$        &\mycc $2.22/0.7$        &\mycc $1.77/0.6$        &\mycc $1.96/0.7$        &\mycc $1.81/0.6$        &\mycc $1.68/0.5$        &\mycc $1.83/0.6$\\
 &
  \mycc  \textbf{SLAM/Radar} &\mycc MAROAM-Full~\cite{MAROAM} &\mycc ~\cite{MAROAM} &\mycc $1.63/0.46$        &\mycc $1.83/0.56$        &\mycc $1.49/0.47$        &\mycc $1.54/0.47$        &\mycc $1.61/0.50$        &\mycc $1.55/0.53$        &\mycc $1.78/0.54$        &\mycc $1.55/0.50$        &\mycc $1.62/0.50$\\


&\mycc  \textbf{SLAM/Radar} &\mycc \tbv{}-T.8 (ours) &\mycc   &\mycc \underbar{1.17/0.35}  &\mycc \underbar{1.15/0.35}  &\mycc \textbf{1.06/0.35}  &\mycc \underbar{1.12}/\textbf{0.37}  &\mycc \underbar{1.09}/\textbf{0.36}  &\mycc \underbar{1.18}/\textbf{0.35}  &\mycc \underbar{1.32}/\textbf{0.36}  &\mycc \textbf{1.10/0.36}  &\mycc \underbar{1.15}/\textbf{0.36} \\
\end{tabular}
\end{adjustbox}
\vspace{-0.2cm}
\caption{ 
Top: ATE [m], Bottom: Drift (\% translation error / deg/$100$~m) on the \texttt{Oxford} dataset~\cite{RadarRobotCarDatasetICRA2020}. Methods marked with p finished prematurely. Methods for Radar SLAM are shaded. The best and second best radar results are \textbf{bold}/\underbar{underlined}.
}\label{tab:TabOxfordATE}
\vspace{-0.1cm}
\end{table*} 
\setcounter{figure}{4}
\begin{figure*}[b!]
	\centering
	\newcommand\figsizey{0.18\hsize}
	\newcommand\figsizex{0.18\hsize}
	\newcommand\trimtop{0.5}
	\newcommand\trimbot{2.8cm}
	\captionsetup[subfloat]{farskip=0pt,captionskip=2pt}
	\begin{center}
		\subfloat[\texttt{10-12-32}]{\includegraphics[trim={0.0cm \trimbot{} 0cm \trimtop cm},clip,width=\figsizex,height=\figsizey]{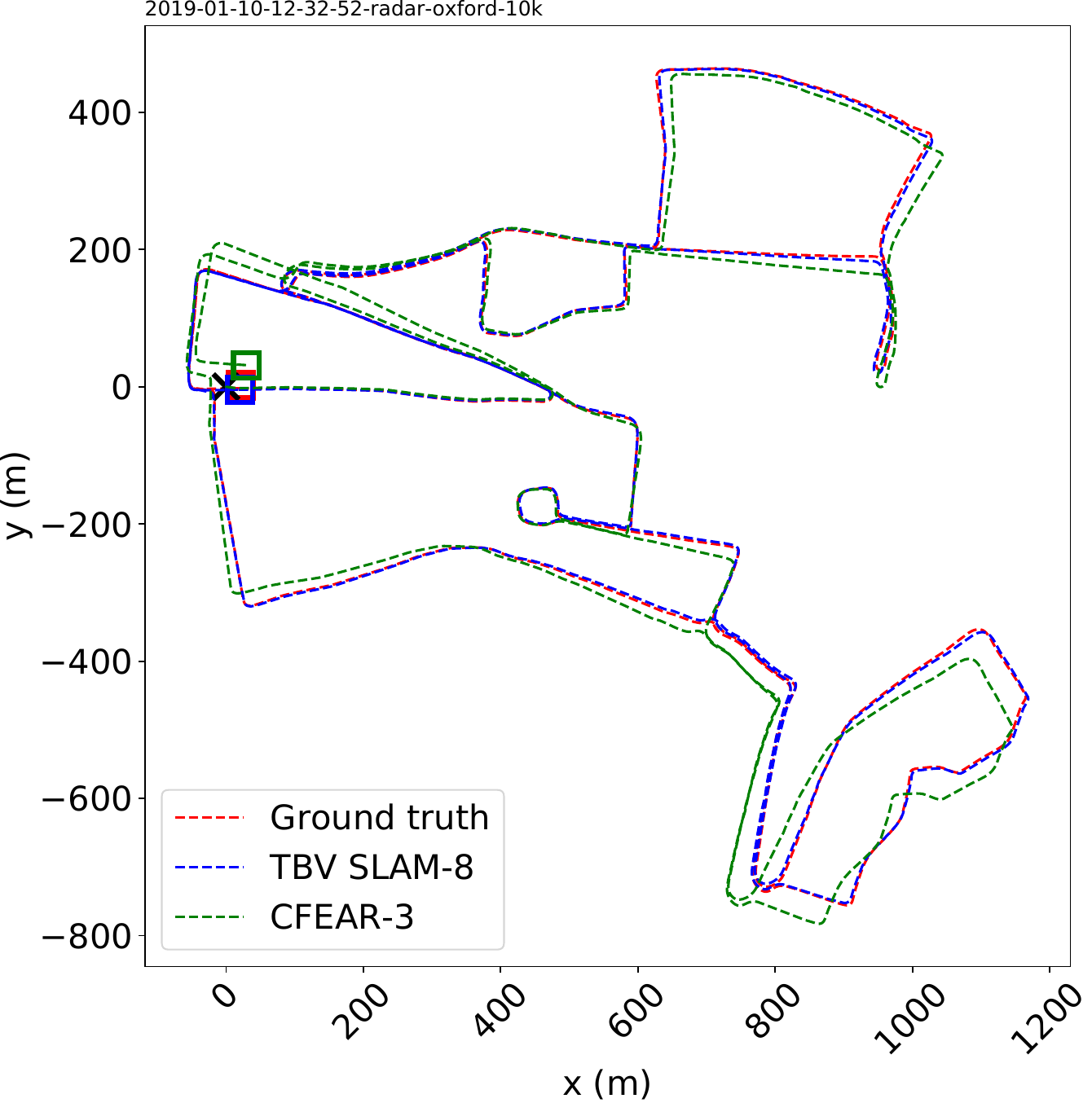}\label{fig:10-12-32}}\hfill
		\subfloat[\texttt{16-13-09}]{\includegraphics[trim={0.0cm \trimbot{} 0cm \trimtop cm},clip,width=\figsizex,height=\figsizey]{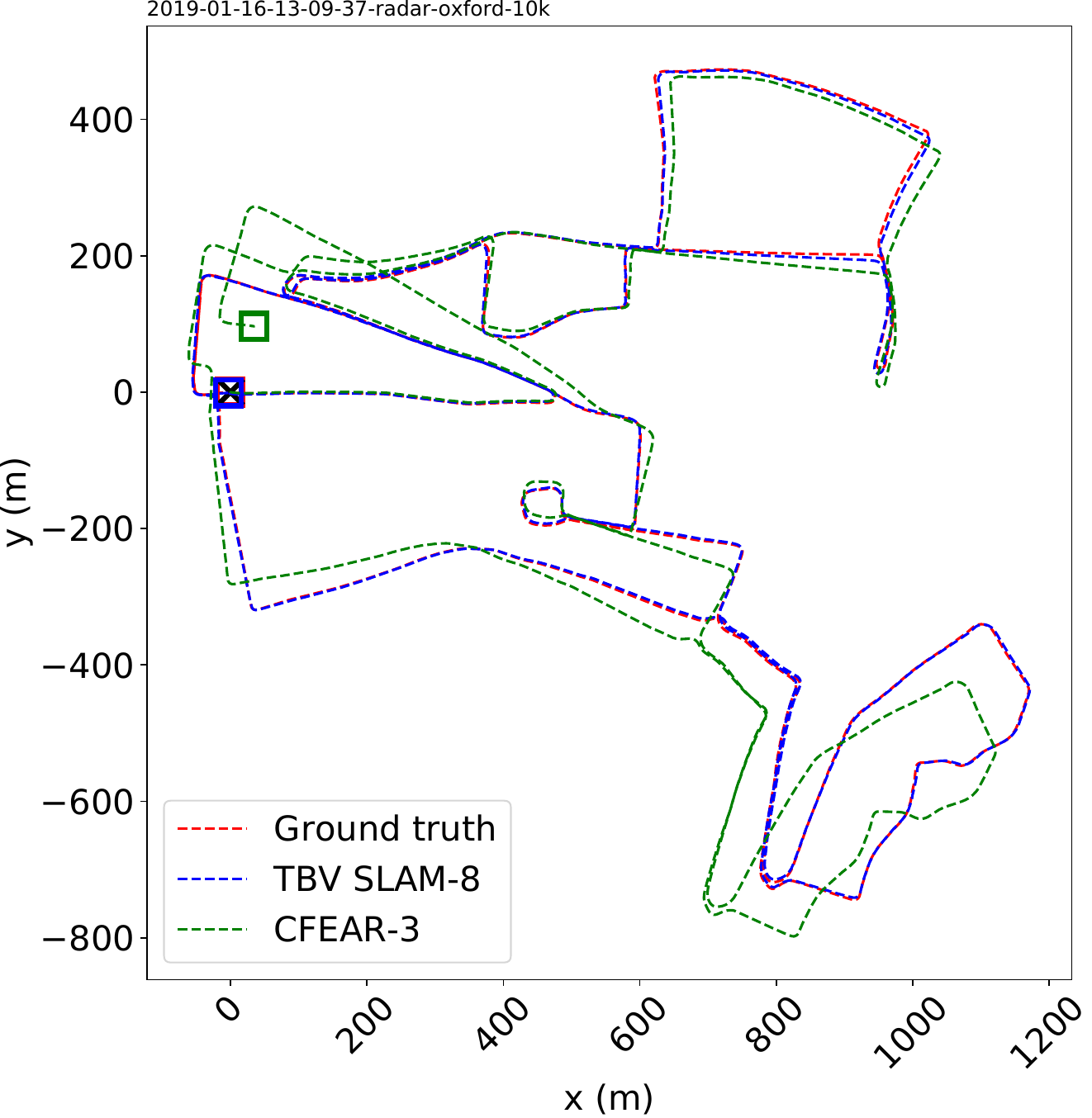}\label{fig:16-13-09}}\hfill
		\subfloat[\texttt{17-13-26}]{\includegraphics[trim={0.0cm \trimbot{} 0cm \trimtop cm},clip,width=\figsizex,height=\figsizey]{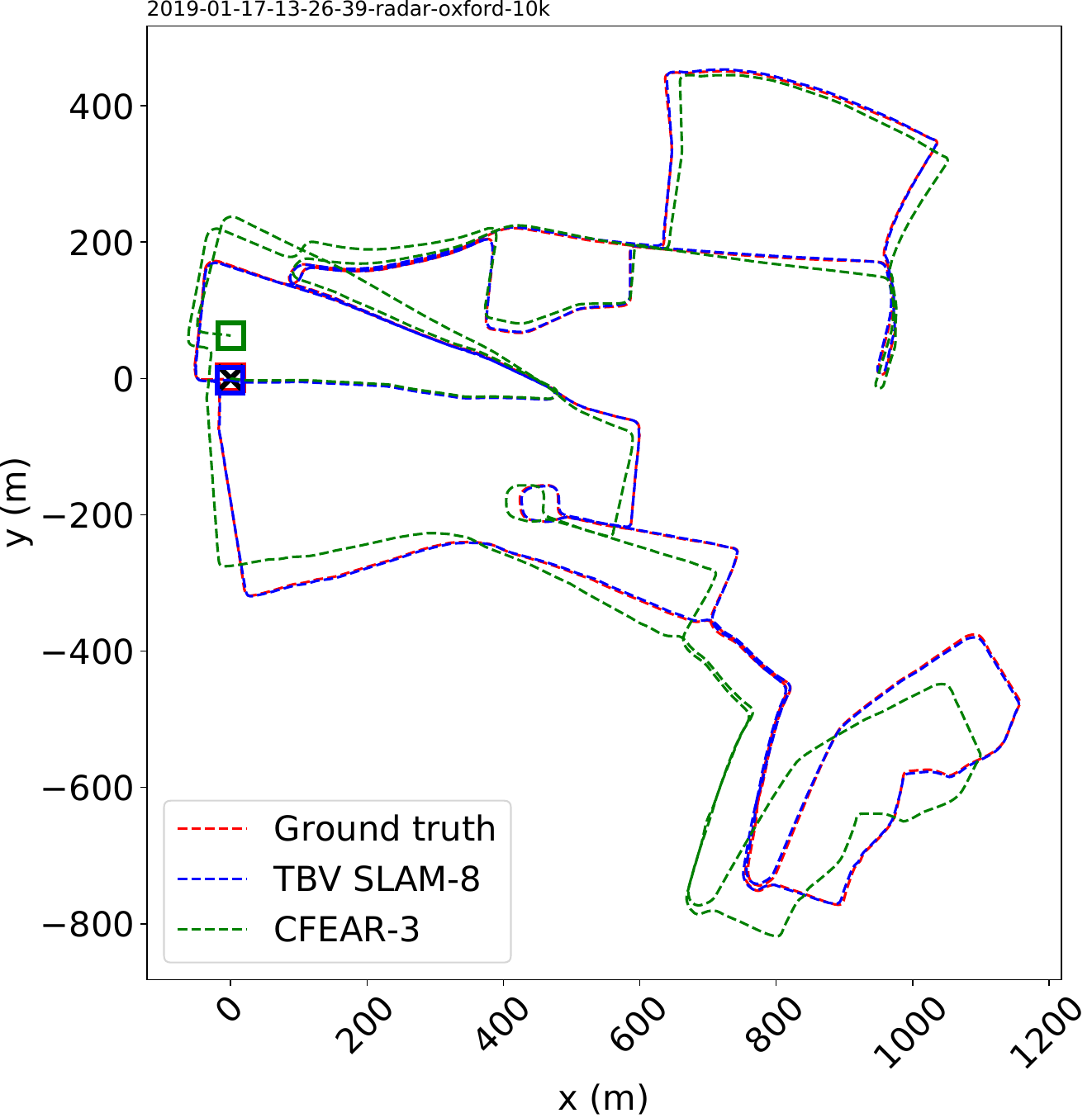}\label{fig:17-13-26}}\hfill
		\subfloat[\texttt{10-11-46}]{\includegraphics[trim={0.0cm \trimbot{} 0cm \trimtop cm},clip,width=\figsizex,height=\figsizey]{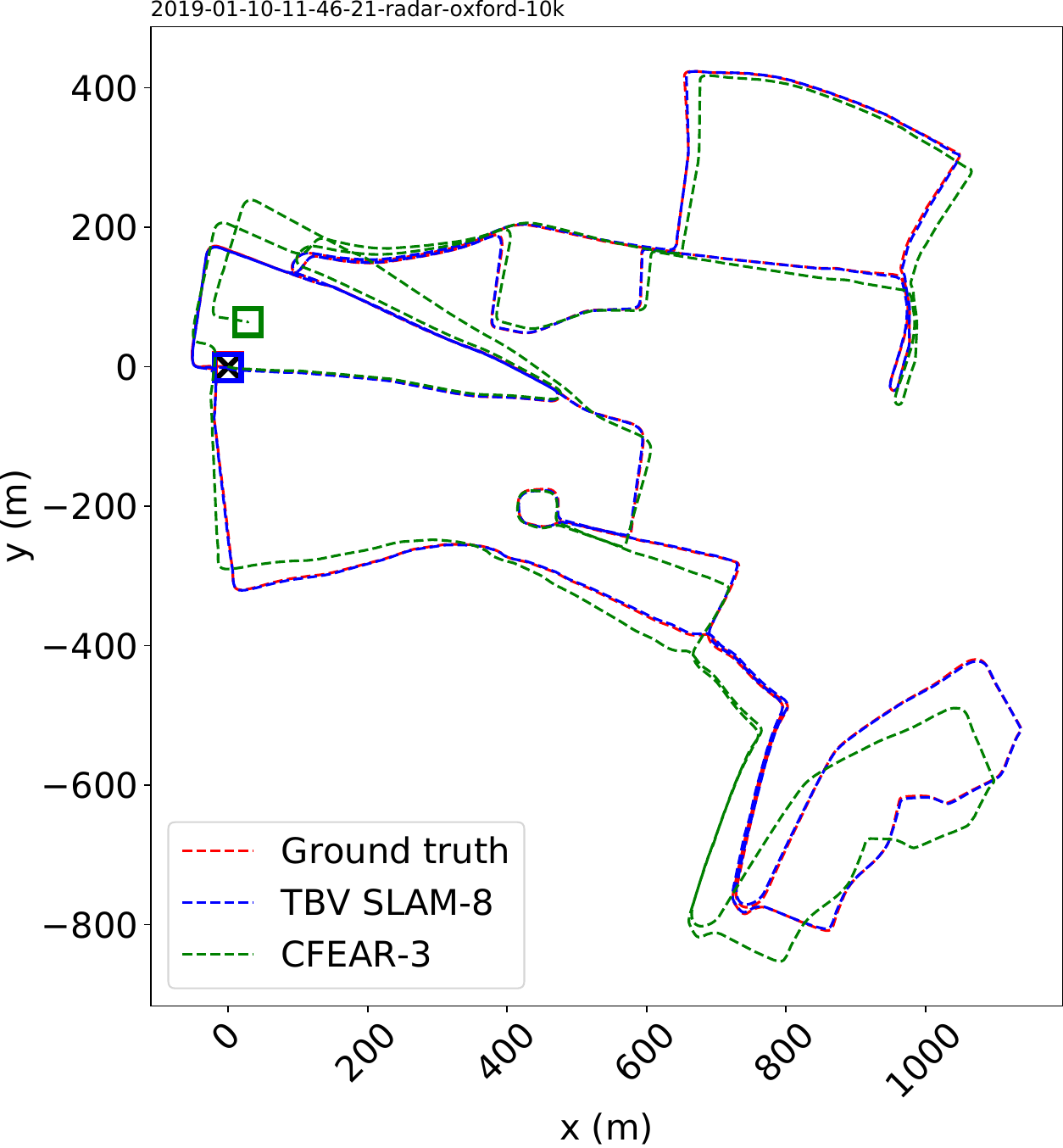}\label{fig:110-11-46}}\hfill\\ 
		\subfloat[\texttt{18-14-14}]{\includegraphics[trim={0.0cm \trimbot{} 0.0cm \trimtop cm},clip,width=\figsizex,height=\figsizey]{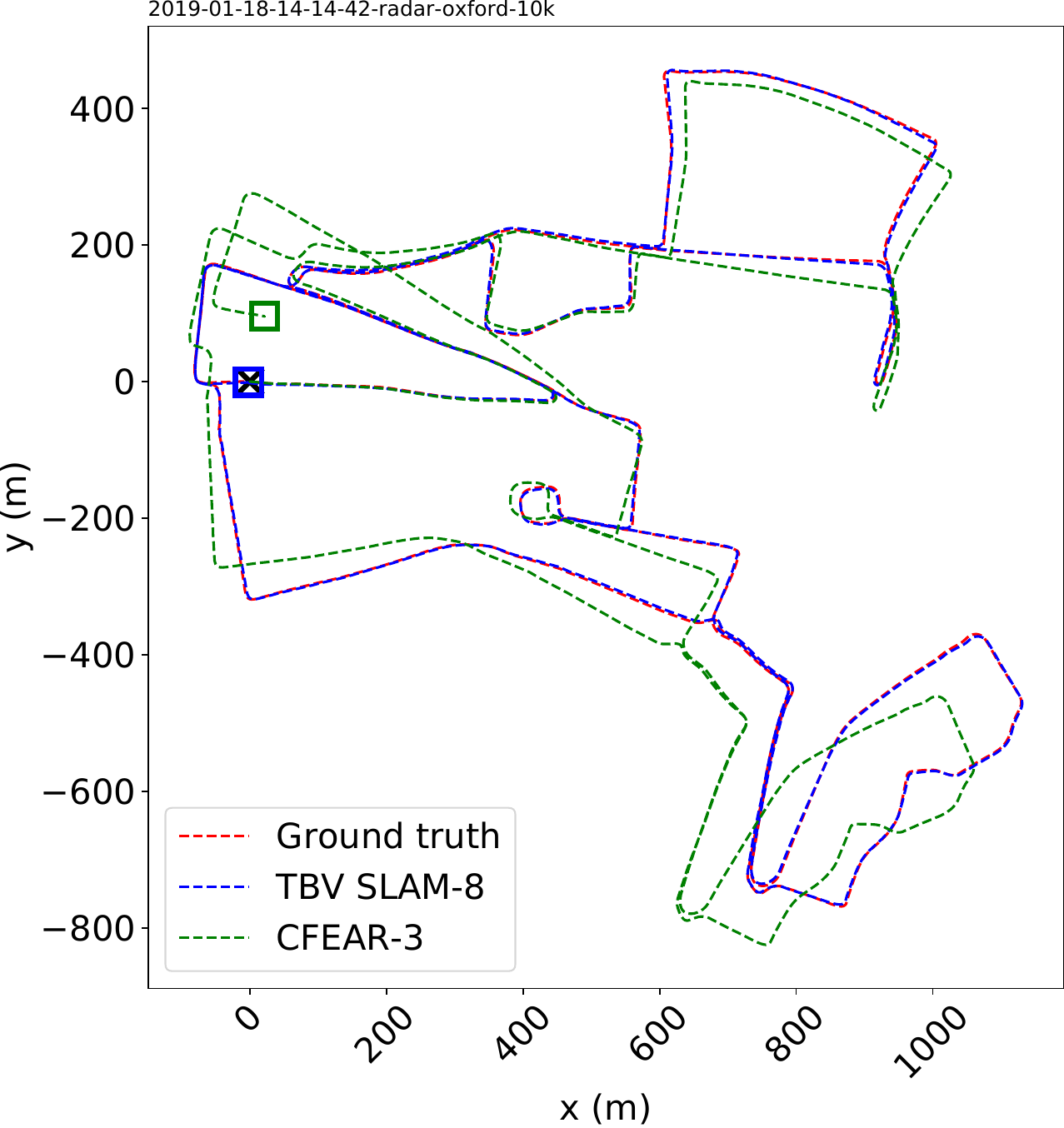}\label{fig:18-14-14}}\hfill
		\subfloat[\texttt{18-15-20}]{\includegraphics[trim={0.0cm \trimbot{} 0cm  \trimtop cm},clip,width=\figsizex,height=\figsizey]{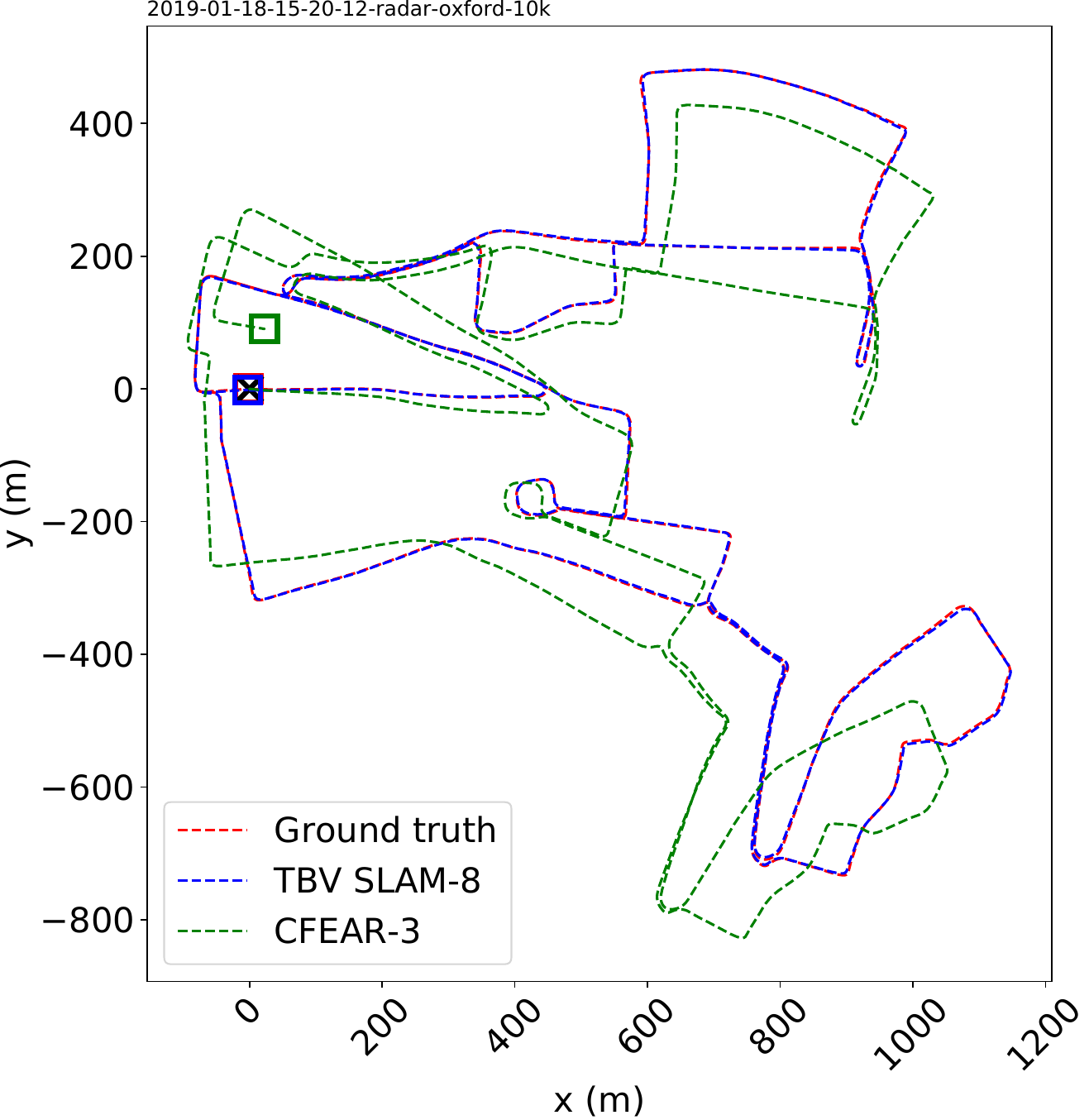}\label{fig:18-15-20}}\hfill
		\subfloat[\texttt{16-11-53}]{\includegraphics[trim={0.0cm \trimbot{} 0cm \trimtop cm},clip,width=\figsizex,height=\figsizey]{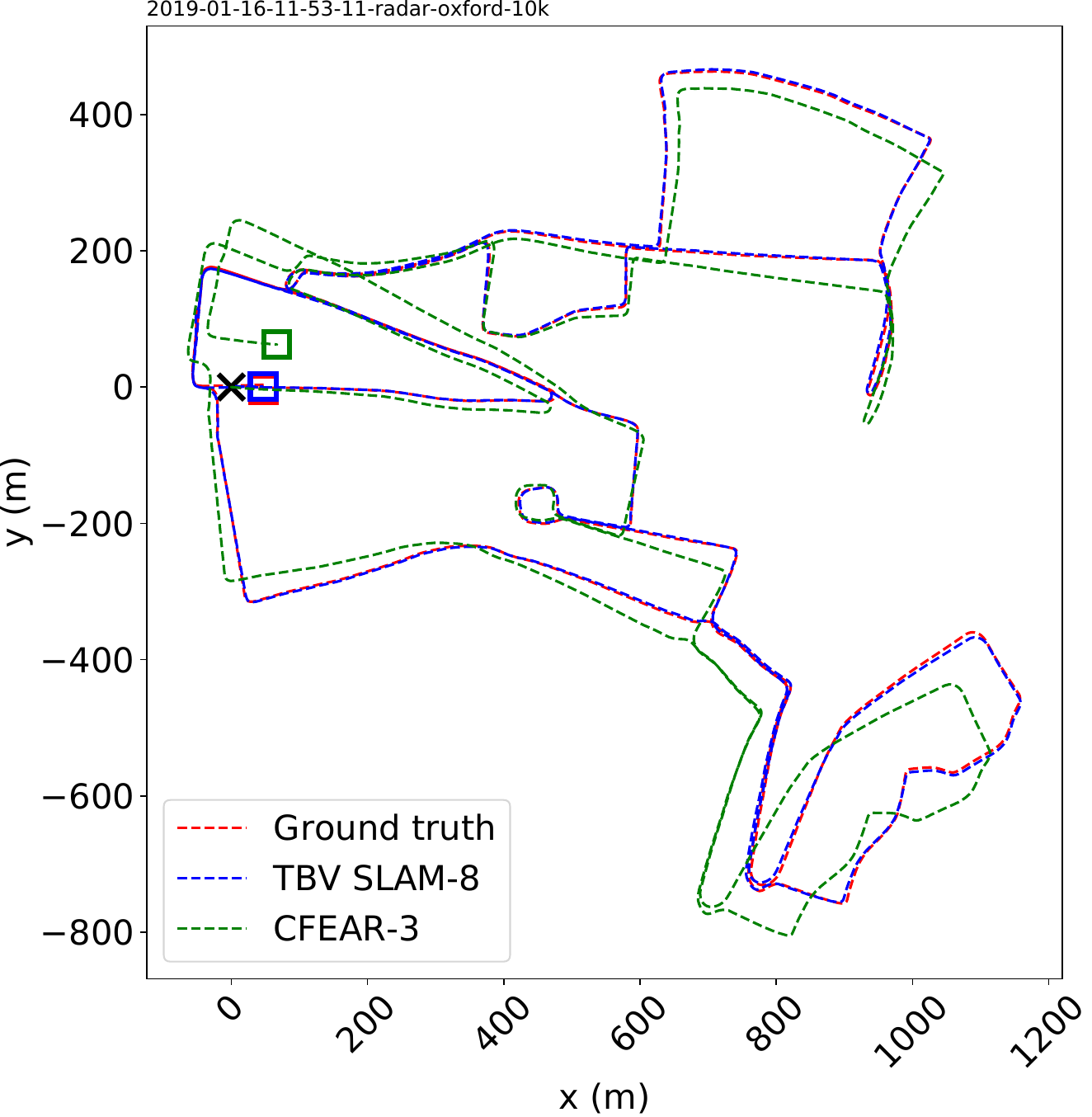}\label{fig:16-11-53}}\hfill
		\subfloat[\texttt{18-14-46}]{\includegraphics[trim={0.0cm \trimbot{} 0cm \trimtop cm},clip,width=\figsizex,height=\figsizey]{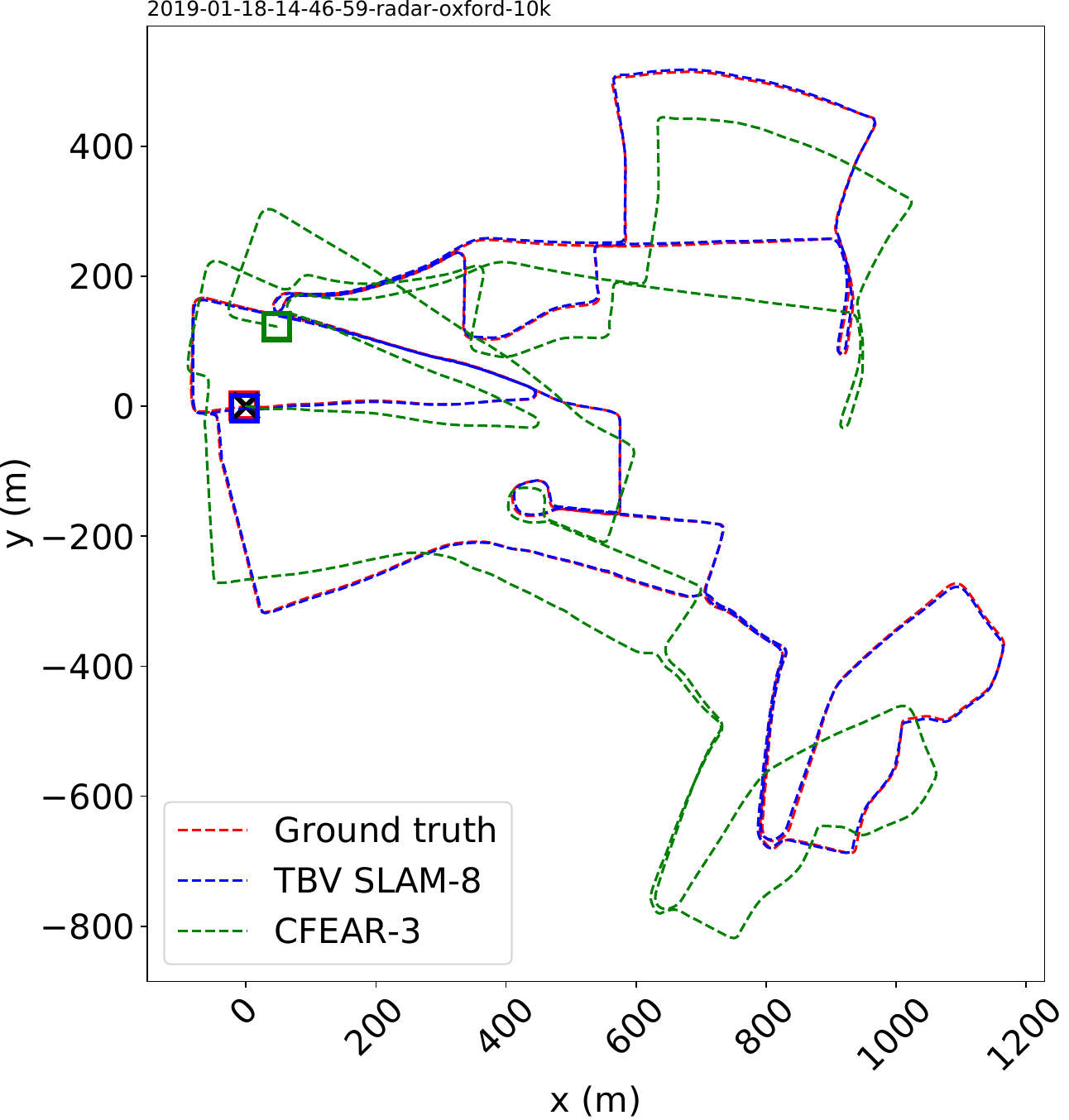}\label{fig:18-14-46}}
  \vspace{-0.25cm}
		\caption{\texttt{Oxford} trajectories using the proposed method {\color{blue} \tbv{}}, compared with {\color{ForestGreen} CFEAR-3~\cite{cfear_journal} (odometry only)} and {\color{red}Ground truth}. 
  \changed{First ($\times$) and final($\square$) positions are marked. Trajectories can be directly compared to} ~\cite{hong2020radarslam,cfear_journal}.\label{fig:sequences_oxford}}
	\end{center}
 \vspace{0cm}
\end{figure*}
\begin{enumerate}[label={T.\arabic*}]
    \item \textit{Radar Scan Context}: Polar radar data is, without preprocessing, directly downsampled into a Scan Context descriptor
    ~\cite{gskim_2020_mulran}. We found the OpenCV interpolation option \texttt{INTER\_AREA} to yield the best results. Verification includes only place similarity $d_{sc}$.\label{para1}
    \item \textit{Aggregated point cloud map}: We instead create the Scan Context descriptor by aggregating motion-compensated peak detections from multiple scans (Sec.~\ref{sec:descriptor_generation}).\label{para2}
    \item \textit{Origin augmentation}: Additional augmentations (origin-shifted copies with lateral offsets) are matched. Out of these, the best match is selected
    (Sec.~\ref{sec:descriptor_generation}).\label{para3}
    \item \textit{Alignment loop verification}: Verification includes alignment quality $d_{align}$ from Sec.~\ref{sec:loop}.\label{para4}
    \item \textit{Odometry decoupled}: Verification includes $d_{odom}$.\label{para5}
    \item \textit{Odometry coupled}: $d_{odom}$ is embedded into the loop retrieval search as described in Sec.~\ref{sec:coupled_matching}.\label{para6}
    \item \textit{Separate verification}: Instead of unified verification, loops are verified separately by alignment ($d_{align}$).\label{para7}
    \item \textit{Multiple candidate selection}: Based on item \ref{para6}. $N_{cand}=3$ competing candidates $\{ (q,c_k)\}$ are used (previously 1). The most likely loop ($y_{loop}^{q,c_k}$) is selected.\label{para8}
\end{enumerate}

The impact is presented in Fig.~\ref{fig:oxford_loop_closure_ablation} for the eight \texttt{Oxford} sequences, and Fig.~\ref{fig:mulran_loop_closure_ablation} for the nine \texttt{MulRan} sequences. Loop detections are visualized in Fig.~\ref{fig:qualitative_loops} for the \texttt{Oxford} sequence \texttt{16-13-09} with $y_{th}=0.9$.
The raw Scan Context (~\ref{para1}) achieves higher recall compared to the sparse local map (\ref{para2}). The difference is larger in \texttt{MulRan}, where scans are acquired in the same moving direction, and motion distortion is less of a challenge. Additionally, we noted that the difference is highest within the feature-poor \texttt{Riverside} sequences. This suggests that maintaining information-rich radar data is advantageous compared to using a sparse local map, especially when features are scarce. Note however that our local mapping technique is primarily motivated by the need for a sparse Cartesian point cloud for efficient augmentation.

\begin{table*}[t!]
\centering
\vspace{0.2cm}
 \begin{adjustbox}{width=\textwidth}
\begin{tabular}{p{3ex}l|l|l|lllllllll|ll}
\cline{2-14}
&\textbf{Type/Modality}& \textbf{Method}  & \textbf{Eval.} & \texttt{KAIST01} & \texttt{KAIST02} & \texttt{KAIST03} & \texttt{DCC01} & \texttt{DCC02} & \texttt{DCC03} & \texttt{RIV.01} & \texttt{RIV.02} & \texttt{RIV.03} & Mean\\
\cline{2-14}
\ldelim\{{7.5}{30pt}[\rotatebox{90}{\hspace{-2mm}{ATE}}]&&&&&&&&&&&&&&  
\\[-2ex]
&SLAM/Lidar & SuMa Full~\cite{behley2018rss}  &~\cite{hongSLAM_IJRR}  &  {38.70} & {31.90} & {46.00} & {13.50} & {17.80} & {29.60} & {-} & {-} & {-} & {22.90}  \\
&Odometry/Lidar&KISS-ICP (odometry)~\cite{vizzo2022arxiv}  &  &  {17.40} & {17.40} & {17.40} & {15.16 } & {15.16} & {15.16} & {49.02} & {49.02} & {49.02} & {27.2}\\
\cline{2-14}
\\[-2ex]
&Odometry/Radar&CFEAR-3~\cite{cfear_journal} (odometry used) & & 7.53 & 7.58 & 12.21 & 6.39 & 3.67 & 5.40 & 6.45 & 3.87 & 19.44 & 8.06\\
&\mycc  \textbf{SLAM/Radar} &\mycc RadarSLAM Full~\cite{hongSLAM_ARXIV}  &\mycc ~\cite{hongSLAM_IJRR} &\mycc {6.90} &\mycc {6.00} &\mycc {4.20} &\mycc {12.90} &\mycc {9.90} &\mycc {3.90} &\mycc {9.00} &\mycc {7.00} &\mycc {10.70} &\mycc {7.80} \\
&\mycc  \textbf{SLAM/Radar} &\mycc MAROAM Full~\cite{MAROAM}  &\mycc ~\cite{MAROAM} &\mycc {-} &\mycc {-} &\mycc {-} &\mycc {-} &\mycc {5.81} &\mycc {-} &\mycc {-} &\mycc {4.85} &\mycc {-} &\mycc {-} \\

&\mycc \textbf{SLAM/Radar} &\mycc TBV SLAM-T.8-dyn-cov (ours) &\mycc &\mycc  \underbar{1.71} &\mycc      \underbar{1.42} &\mycc      \underbar{1.52} &\mycc    \textbf{5.41} &\mycc    \textbf{3.29} &\mycc    \textbf{2.61} &\mycc  \underbar{2.66} &\mycc          \underbar{2.49} &\mycc \underbar{2.52} &\mycc   \underbar{2.63}  \\
&\mycc \textbf{SLAM/Radar} &\mycc TBV SLAM-T.8-cov (ours) &\mycc &\mycc \textbf{1.66} &\mycc \textbf{1.39} &\mycc \textbf{1.50}  &\mycc \underbar{5.44} &\mycc \underbar{3.32} &\mycc \underbar{2.66} &\mycc \textbf{2.61} &\mycc \textbf{2.36} &\mycc \textbf{1.48} &\mycc \textbf{2.49} \\
\cline{2-14}\\[-2ex]
\cline{2-14} 
\ldelim\{{6.5}{30pt}[\rotatebox{90}{\hspace{-2mm}{Drift}}]&&&&&&&&&&&&&&  
\\[-2ex]
&SLAM/Lidar & SuMa Full~\cite{behley2018rss}  &   & 2.9/0.8 & 2.64/0.6 & 2.17/0.6 & 2.71/0.4 & 4.07/0.9 & 2.14/0.6 & 1.66/0.6$^P$ & 1.49/0.5$^P$ & 1.65/0.4$^P$ & 2.38/0.5 & \\
&Odometry/Lidar & KISS-ICP (odometry)~\cite{vizzo2022arxiv}  &  &  {2.28/0.68} & {2.28/0.68} & {2.28/0.68} & {2.34/0.64} & {2.34/0.64 } & {2.34/0.64 } & {2.89/0.64} & {2.89/0.64} & {2.89/0.64} & {2.5/0.65} &\\
\cline{2-14} \\[-2ex]
&Odometry/Radar & CFEAR-3-s50~\cite{cfear_journal}  & ~\cite{cfear_journal}  &  \underbar{1.48}/0.65 & \underbar{1.51}/0.63 & \underbar{1.59}/0.75 & \underbar{2.09}/0.55 & \underbar{1.38}/0.47 & \underbar{1.26}/0.47 & 1.62/\underbar{0.62} & \underbar{1.35}/0.52 & 
\underbar{1.19/0.37} & \underbar{1.50}/0.56 &
  
\\
&Odometry/Radar&CFEAR-3~\cite{cfear_journal} (odometry used)  & & 1.59/0.66 & 1.62/0.66 & 1.73/0.78 & 2.28/0.54 & 1.49/0.46 & 1.47/0.48 & \underbar{1.59}/0.63 & 1.39/\underbar{0.51} & 1.41/0.40 &  1.62/0.57 &
\\
&\mycc   \textbf{SLAM/Radar} &\mycc RadarSLAM-Full~\cite{hongSLAM_ARXIV}  &\mycc ~\cite{hongSLAM_IJRR} &\mycc  1.75/\underbar{0.5} &\mycc 1.76/\underbar{0.4} &\mycc 1.72/\underbar{0.4} &\mycc 2.39/\underbar{0.4} &\mycc 1.90/\underbar{0.4} &\mycc 1.56/\textbf{0.2} &\mycc  3.40/0.9 &\mycc 1.79/\textbf{0.3} &\mycc 1.95/0.5 &\mycc 2.02/\underbar{0.4} \\
&\mycc  \textbf{SLAM/Radar} &\mycc  TBV SLAM-T.8 (ours) &\mycc  &\mycc \textbf{1.01/0.30} &\mycc \textbf{1.03/0.30} &\mycc \textbf{1.08/0.35} &\mycc \textbf{2.01/0.27} &\mycc \textbf{1.35/0.25} &\mycc \textbf{1.14}/\underbar{0.22} &\mycc \textbf{1.25/0.35} &\mycc \textbf{1.09/0.30} &\mycc \textbf{0.99/0.18} &\mycc \textbf{1.22/0.28} \\
\cline{2-14} 
\end{tabular}
  \end{adjustbox}
\caption{ Top: Absolute Trajectory error (ATE), Bottom: Drift (\% translation error / deg/$100$~m) on the \texttt{MulRan} dataset~\cite{gskim_2020_mulran}. 
\changed{Methods for Radar SLAM are shaded.
The best and second best radar results are marked in \textbf{bold}/\underbar{underlined}.}
}
\label{tab:TabMulRanATE}
\end{table*}
\newcommand\trimbot{2.9cm}
\newcommand\trimbotalt{2.5cm}
\newcommand\trimtop{0.4cm}
\setcounter{figure}{5}
\begin{figure*}[t!]
\vspace{-0.4cm}
  \begin{center}
    \captionsetup[subfigure]{margin={0cm,0.0cm}}
    \subfloat[\texttt{KAIST01}]{\includegraphics[trim={0.0cm \trimbot{} 0cm \trimtop{}},clip,height=0.18\hsize]{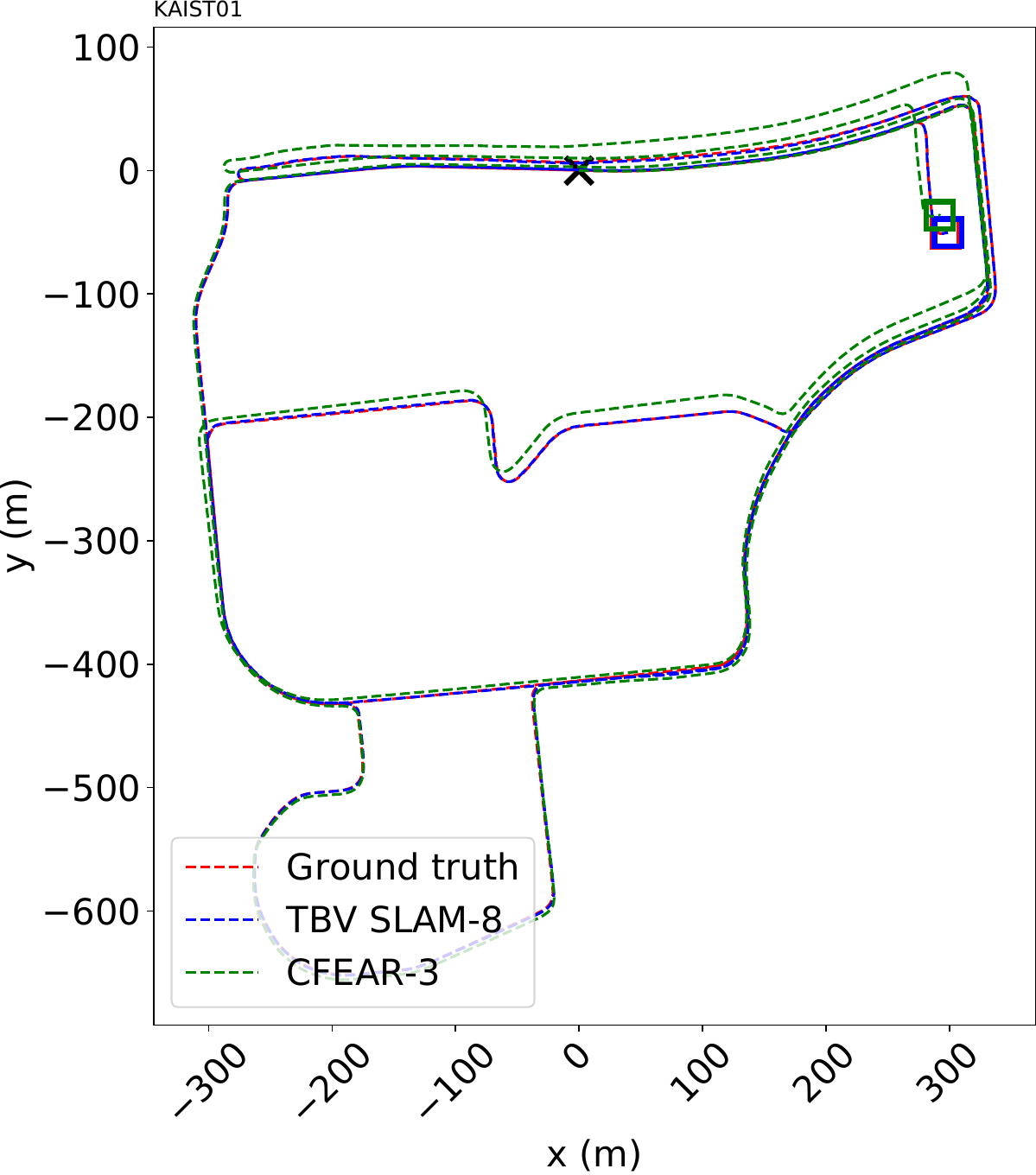}\label{fig:KAIST01}}\hfill
    \subfloat[\texttt{KAIST02}]{\includegraphics[trim={0.0cm \trimbot{} 0cm \trimtop{}},clip,height=0.18\hsize]{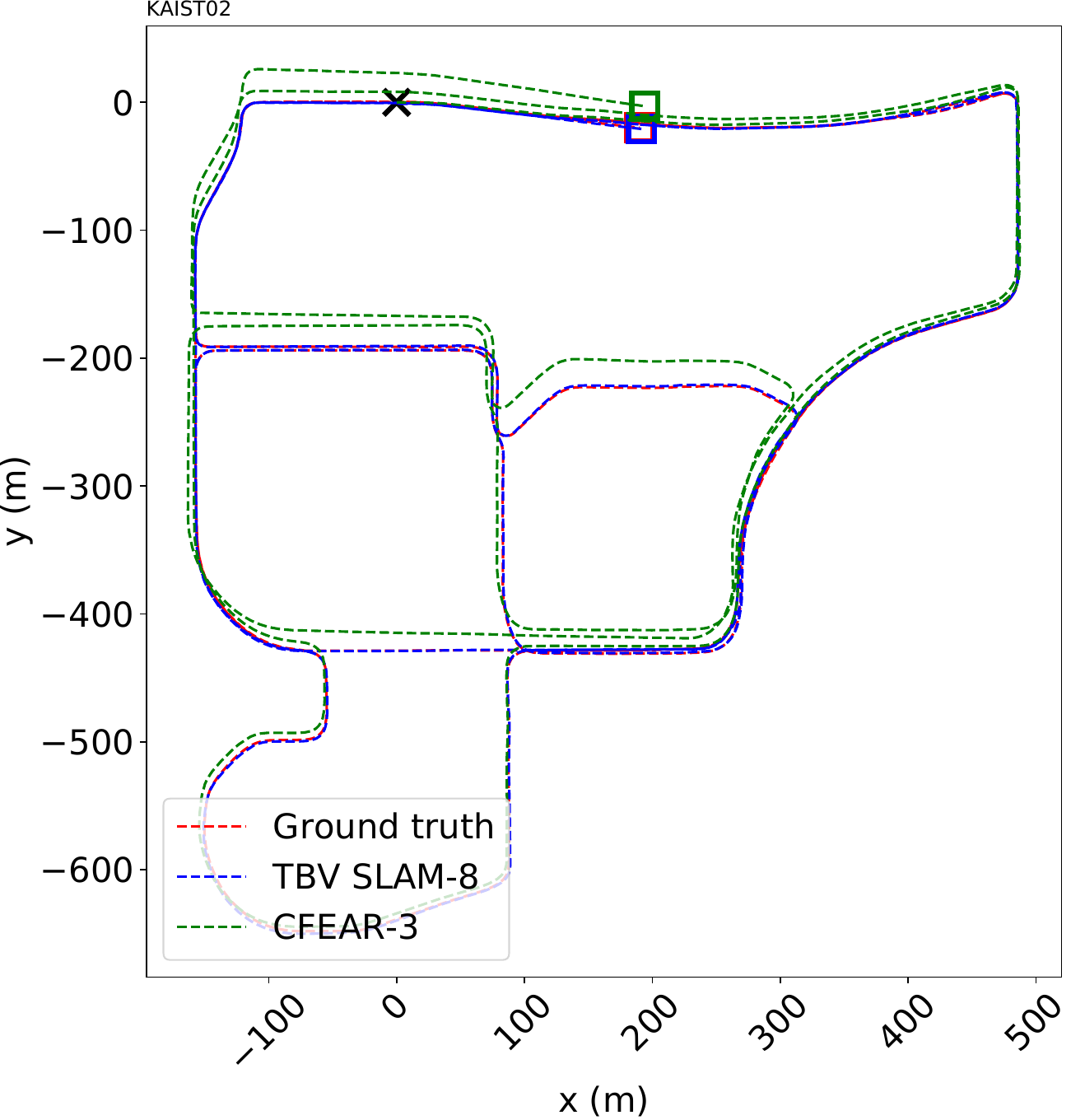}\label{fig:KAIST02}}\hfill
    \subfloat[\texttt{KAIST03}]{\includegraphics[trim={0.0cm \trimbot{} 0cm \trimtop{}},clip,height=0.18\hsize]{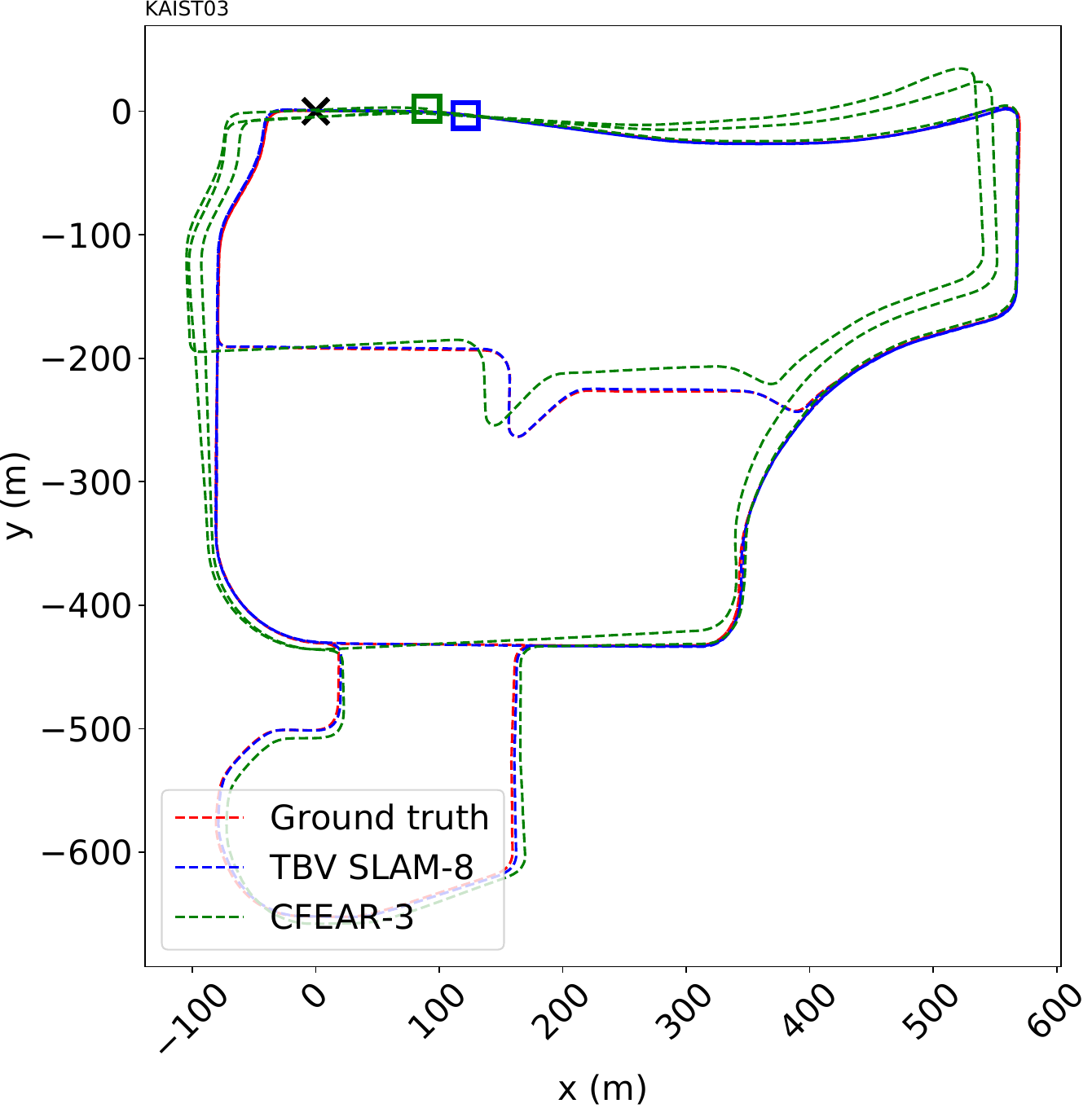}\label{fig:KAIST03}}\hfill
    \subfloat[\texttt{DCC01}]{\includegraphics[trim={0.0cm \trimbot{} 0cm \trimtop{}},clip,height=0.18\hsize]{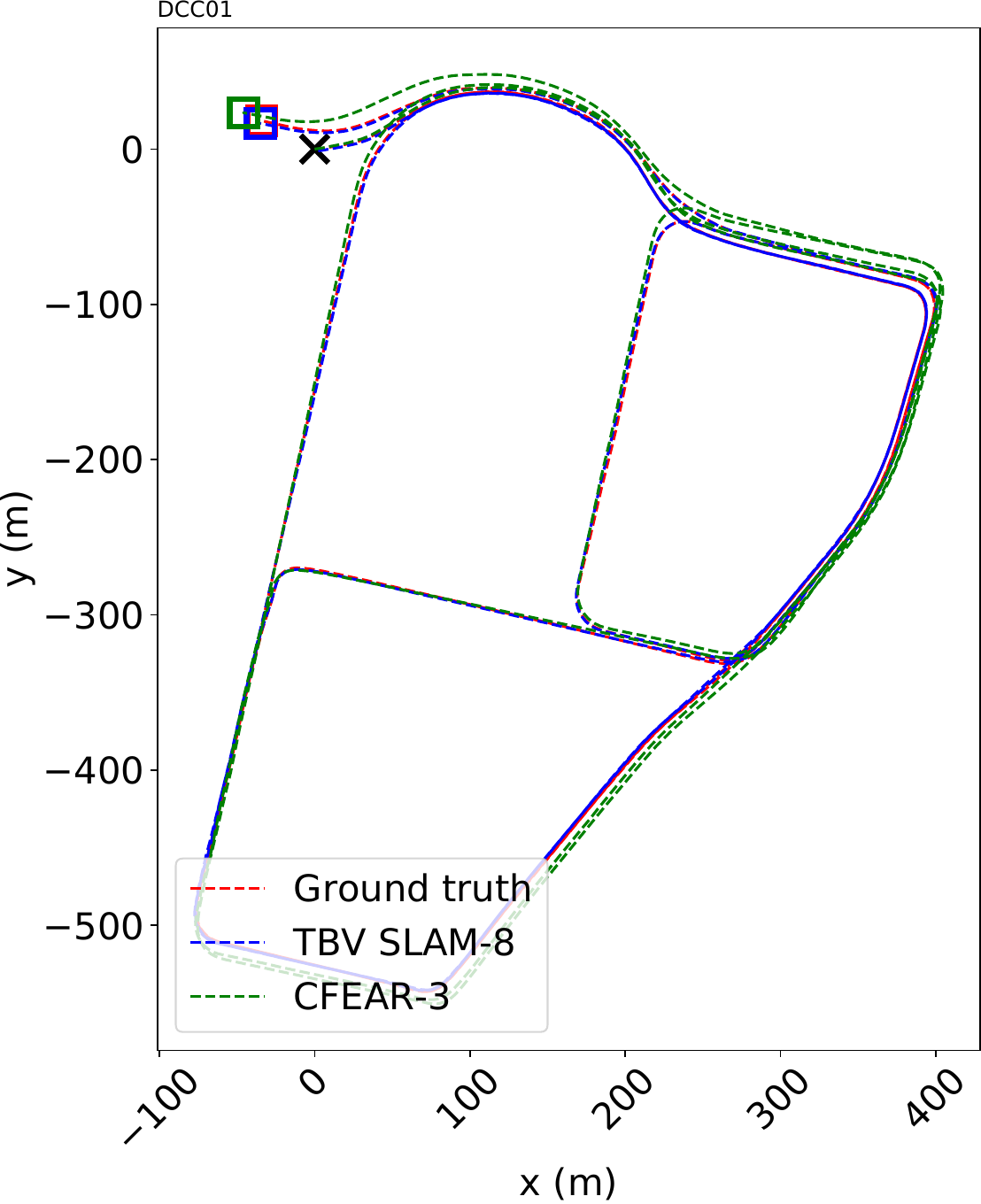}\label{fig:DCC01}}\hfill
    \vspace{-0.3cm}
    \subfloat[\texttt{DCC02}]{\includegraphics[trim={0.0cm \trimbotalt{} 0cm \trimtop{}},clip,height=0.18\hsize]{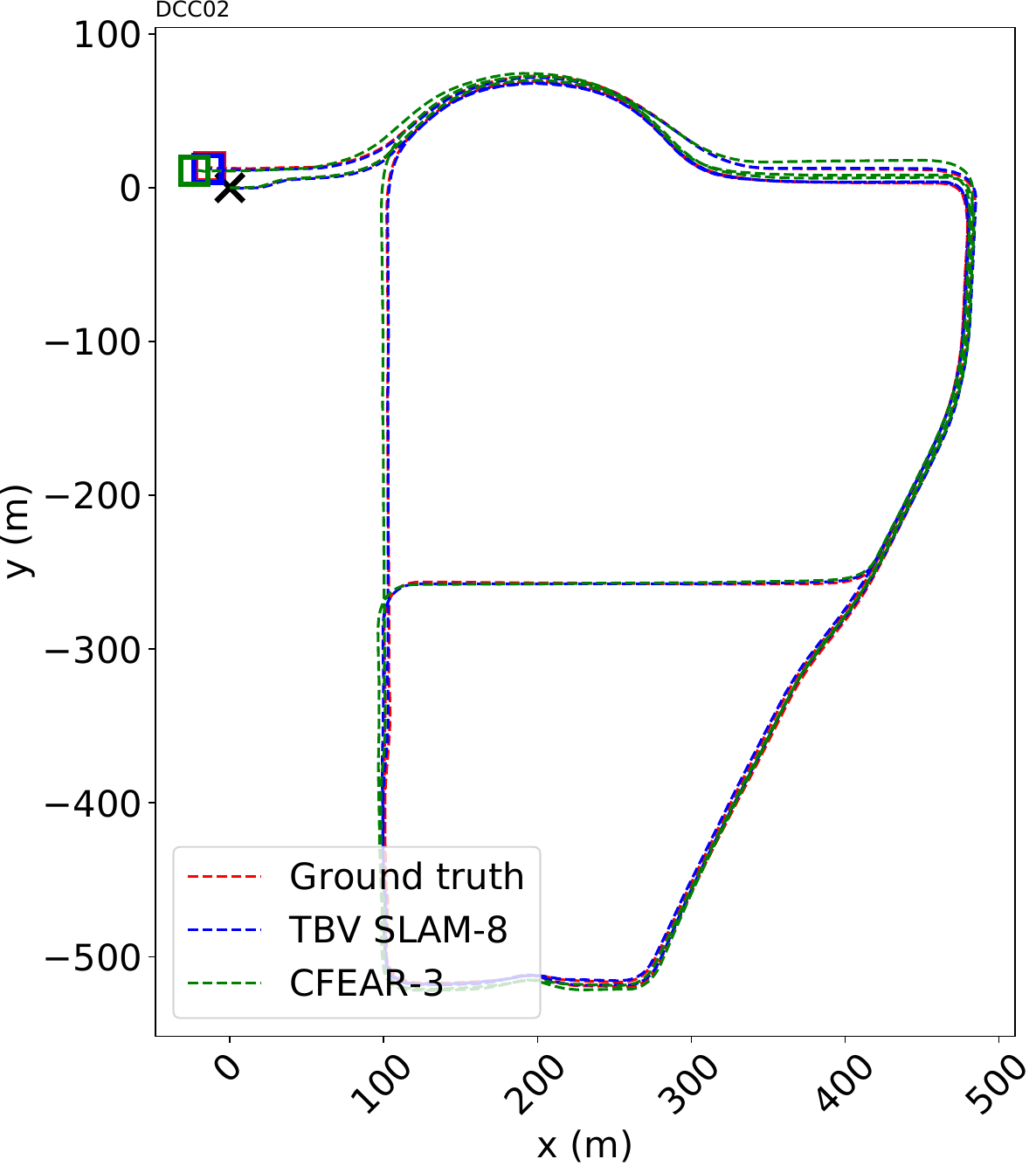}\label{fig:DCC02}}\hfill\\ 
    \subfloat[\texttt{DCC03}]{\includegraphics[trim={0.0cm \trimbotalt{} 0cm \trimtop{}},clip,height=0.18\hsize]{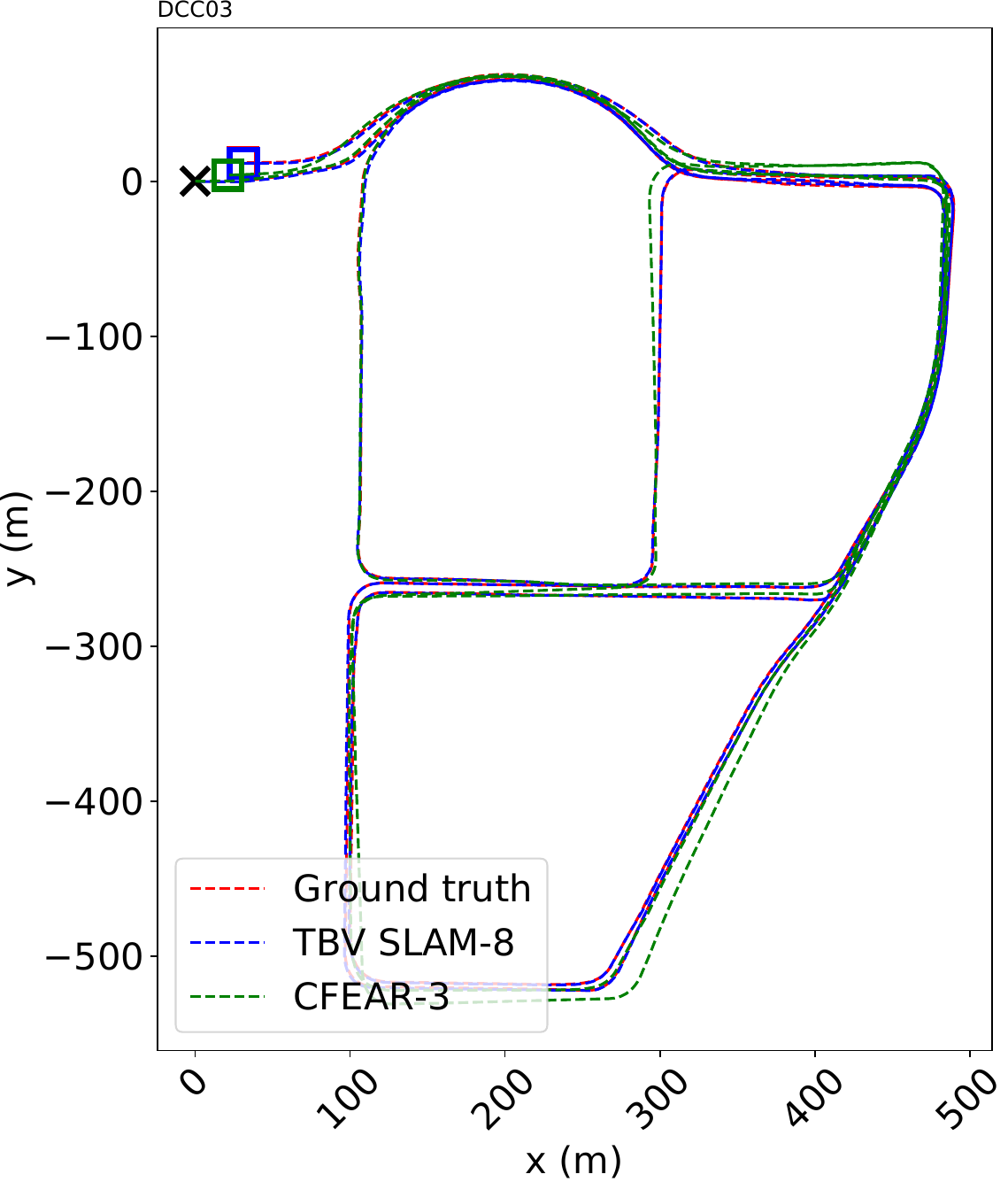}\label{fig:DCC03}}\hfill
    \hspace{1cm}
    \subfloat[][\texttt{RIV.01}]{\includegraphics[trim={0.0cm \trimbotalt{} 0cm 0.5cm},clip,height=0.18\hsize]{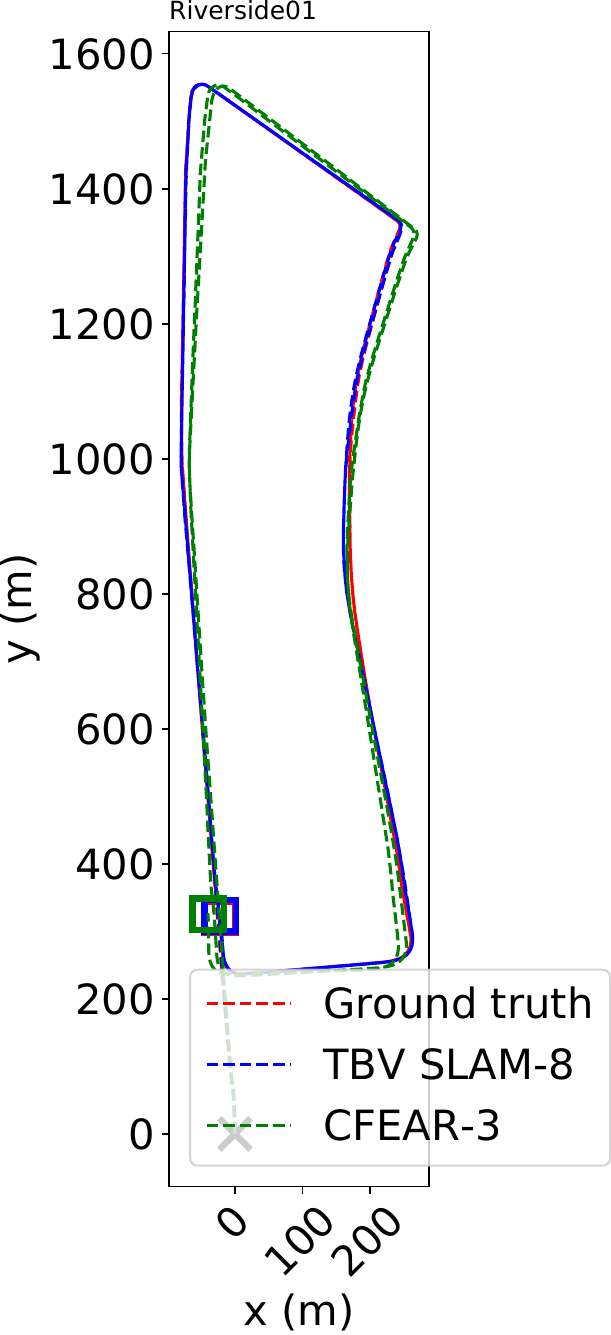}\label{fig:Riverside01}}\hfill
    \subfloat[\texttt{RIV.02}]{\includegraphics[trim={0.0cm \trimbotalt{} 0cm 0.4cm},clip,height=0.18\hsize]{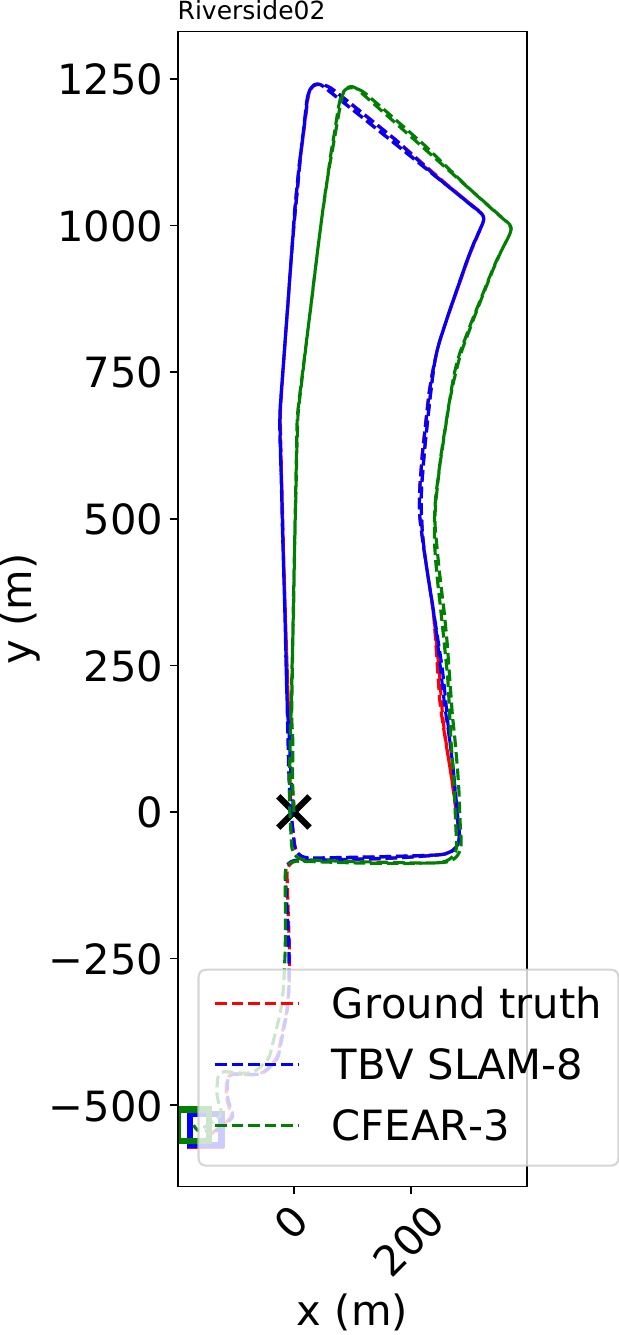}\label{fig:Riverside02}}\hfill
    \subfloat[\texttt{RIV.03}]{\includegraphics[trim={0.0cm \trimbotalt{} 0cm 0.4cm},clip,height=0.18\hsize]{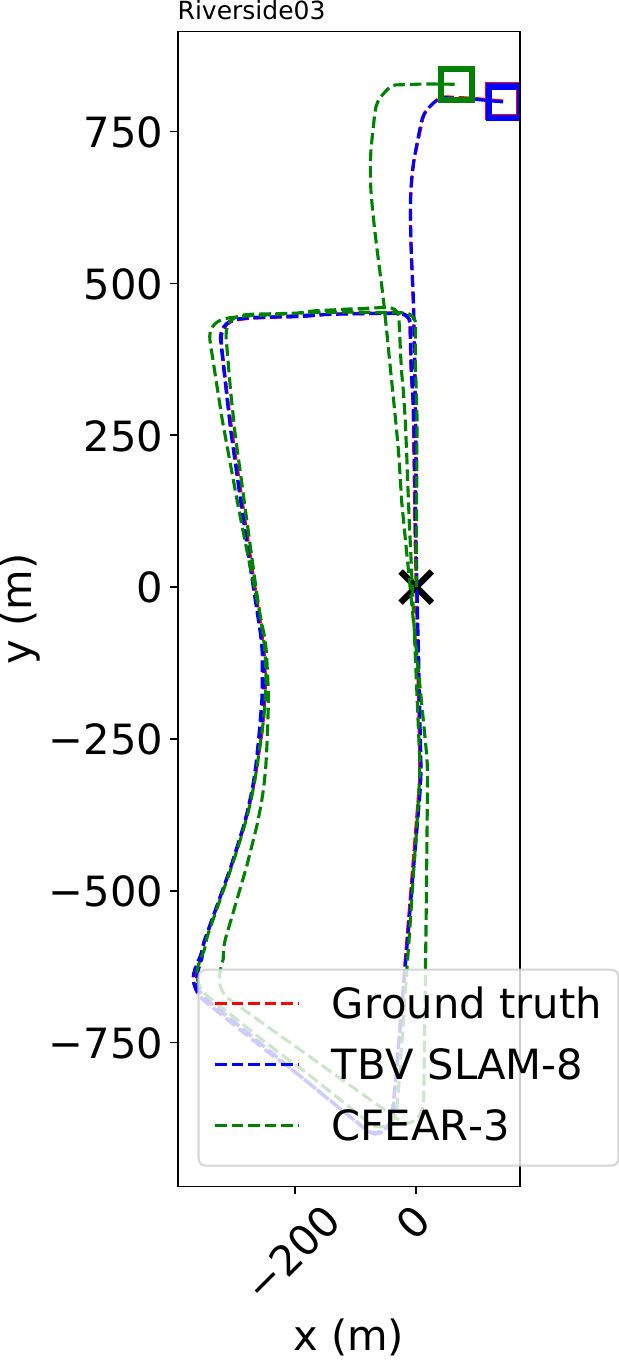}\label{fig:Riverside03}}\hfill
    \subfloat[][\texttt{VolvoCE} - forest]{\includegraphics[trim={2cm 2cm 4cm 3cm},clip,height=0.18\hsize]{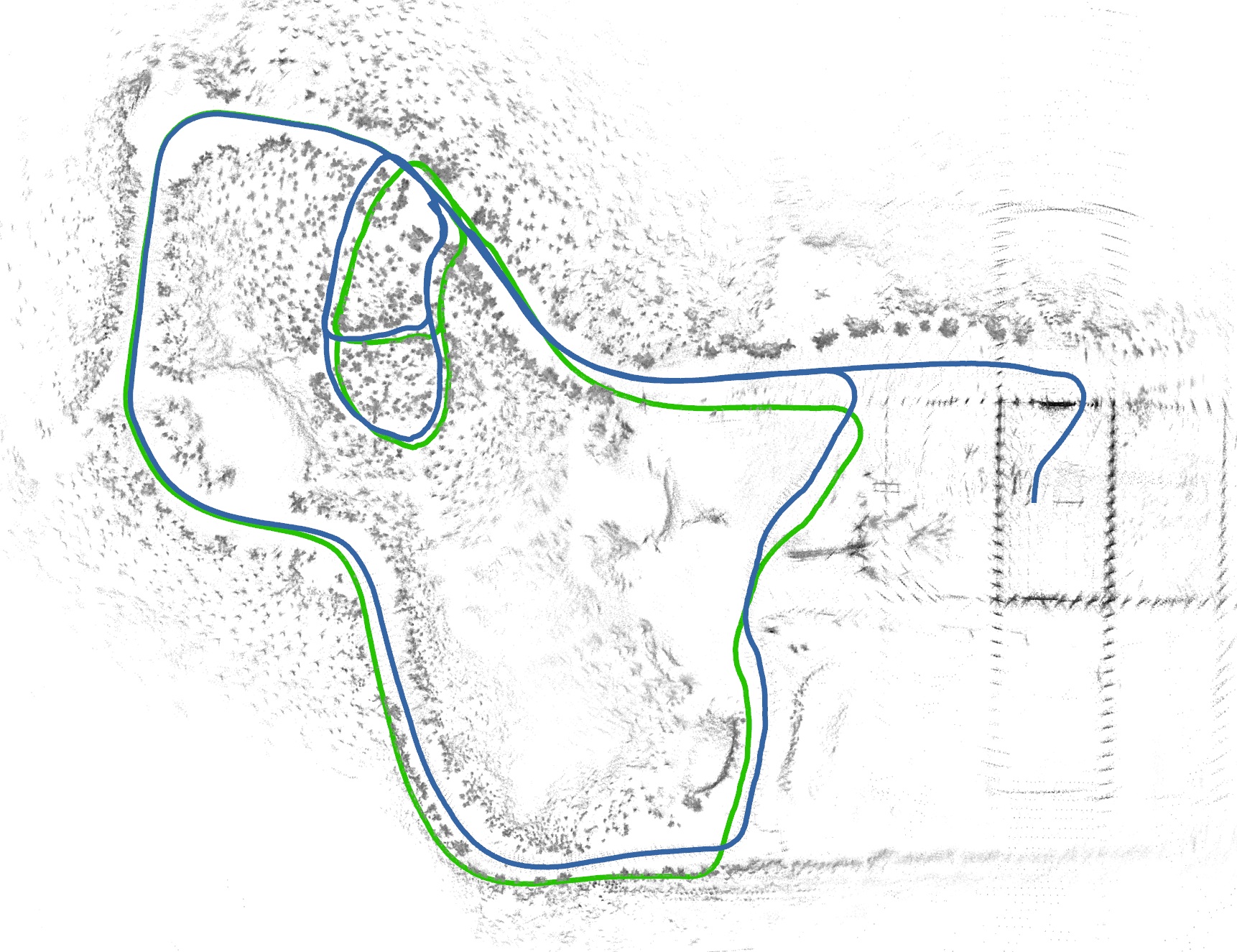}\label{fig:volvo_qual}}\hfill \subfloat[][\texttt{Kvarntorp} - mine]{\includegraphics[trim={3cm 0cm 5cm 1cm},clip,height=0.18\hsize]{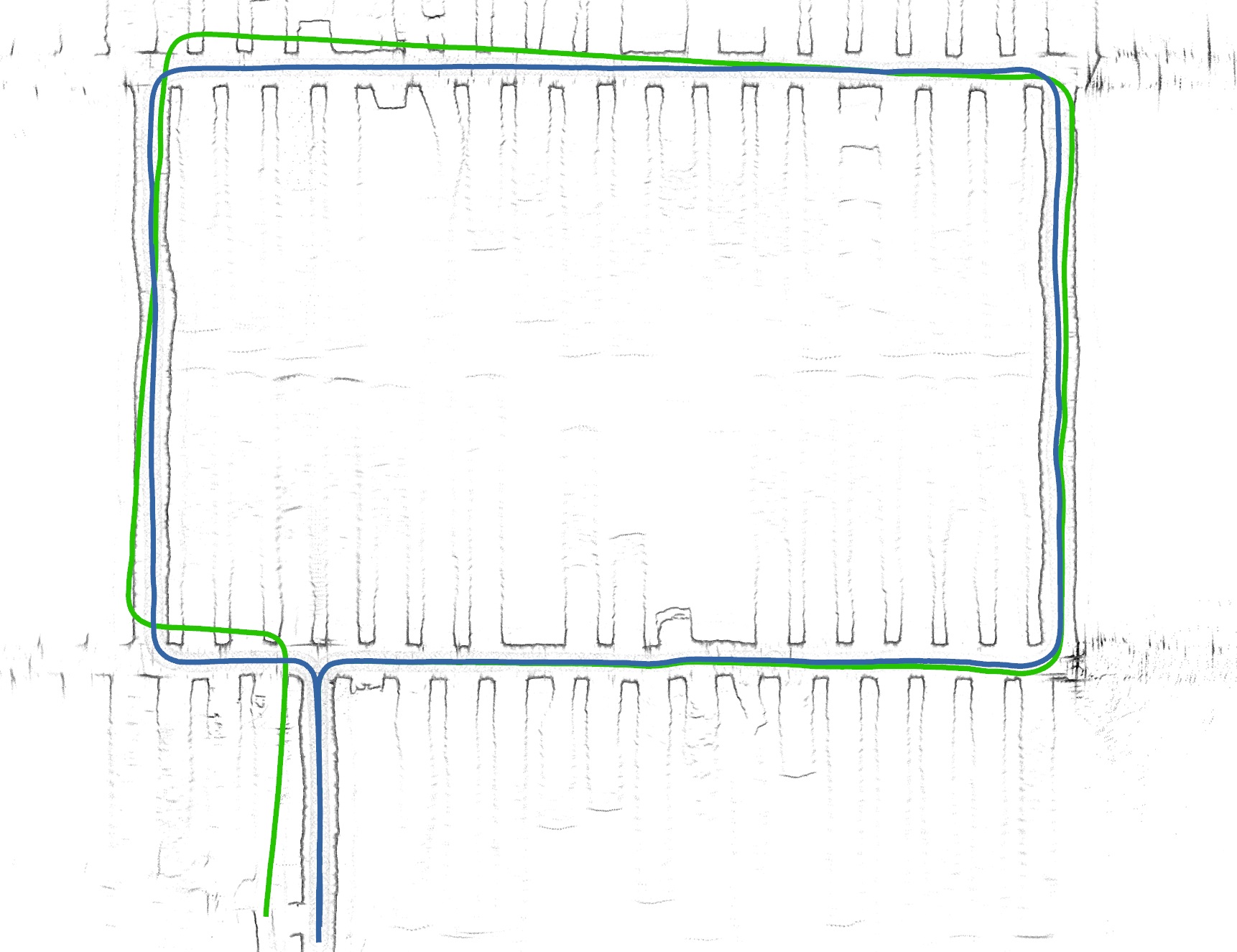}\label{fig:kvarntorp_qual}}
  \end{center}
  \vspace{-0.4cm}
  \caption{ {\color{blue} \tbv{}}, {\color{ForestGreen} CFEAR-3~\cite{cfear_journal} (odometry only)}, and {\color{red}Ground truth}. First ($\times$) and final($\square$) positions are marked.
  \label{fig:MulRanTrajectories}}
  \vspace{-0.6cm}
\end{figure*}

\texttt{Oxford} is more challenging compared to \texttt{MulRan} as a majority of the revisits occur in opposite road lanes and directions.
However, the augmentation technique (\ref{para3}) allows the detection of additional candidates with higher lateral displacement, and as expected, increases the highest top recall, yet at a cost of lower precision. 
This loss of precision can however be alleviated by including \changed{verification based on carefully assessed alignment quality (\ref{para4})}. The improvement is larger in \texttt{Oxford} and we believe the more structured scenes are favorable for alignment analysis.
The decoupled odometry approach (\ref{para5}), which extends verification by including odometry uncertainty, gives a higher tolerance to false positives. \changed{However, coupling odometry uncertainty into the candidate search (\ref{para6}) substantially increases recall to over 90\%, effectively constraining the search space.}
\changed{Verifying candidates jointly with multiple sources (\ref{para6}) circumvents early discarding of candidates based on the mistakenly low confidence, 
thus preferred over cascaded verification (\ref{para7}).
Unified verification and loop retrieval (\ref{para8})--which extends (\ref{para6}) by selecting between multiple candidates--allows for the selection of non-primary retrievals with higher computed alignment quality and confidence. Thus, increasing retrieval and precision in the \texttt{Oxford} dataset.}






\subsection{SLAM performance -- comparative evaluation}

We compare \tbv{} to previous methods for radar, lidar, and visual SLAM within the Oxford and Mulran dataset. 
We primarily compare methods over full trajectories i.e. Absolute Trajectory Error (ATE) $ATE_{RMSE}=\sqrt{\frac{1}{n}\sum_{i=1}^{i=n} ||transl(\mathbf{y}^{est}_i) - transl(\mathbf{y}^{gt}_{i}))||^2}$. Additionally, we provide the KITTI odometry metric~\cite{Geiger2012CVPR}, which computes the relative error between 100-800~m, e.g. error over a shorter distance.
ATE metrics for method MAROAM~\cite{MAROAM} was kindly computed and provided by Wang et al. for this letter.
We tuned our parameters on the \texttt{Oxford} sequence \texttt{10-12-32} and evaluated the performance of SLAM on all other \texttt{Oxford} and \texttt{MulRan} sequences.

The estimated trajectories are depicted in Fig.~\ref{fig:sequences_oxford} and Fig.~\ref{fig:MulRanTrajectories}(a-i). We found that \tbvshort{} effortlessly closes loops and corrects odometry in all sequences. ATE is substantially corrected over the full trajectory, with slightly reduced drift (Tab.~\ref{tab:TabOxfordATE} \& Tab.~\ref{tab:TabMulRanATE}). 
TBV outperforms previous methods for radar SLAM in terms of ATE and drift over all sequences. 

Hence, we conclude that our method improves the state of the art in radar SLAM. 
Surprisingly, we did not observe any improvement from using dynamic covariance (dyn-cov) compared to fixed.
The Hessian-approximated covariance occasionally under- or over-estimates the odometry uncertainty \cite{Landry2019} and thus deteriorates the optimization process. 

\subsection{Generalization to off-road environments}
Finally, we tested \tbvshort{}  on the sequences \texttt{Kvarntorp} and \texttt{VolvoCE} from the \textit{Diverse ORU Radar Dataset~\cite{cfear_journal}}, see footnote for a demo%
\footnote{ORU dataset download: \url{https://tinyurl.com/radarDataset}. 
Demo video:  \url{https://tinyurl.com/TBV-KvarntorpVolvo}.}.
\texttt{Kvarntorp} is an underground mine with partly \changed{feature-poor} sections, while \texttt{VolvoCE} is a mixed environment with forest and open fields.
Trajectories are visualized in Fig.~\ref{fig:MulRanTrajectories}.(j-k). We found that \tbvshort{} was able to produce smooth and globally consistent maps, 
through substantially different environments, 
including challenging road conditions -- without any parameter changes.



\section{Conclusions}
We proposed \tbv{} -- a method for robust and accurate large-scale SLAM using a spinning 2D radar.
We showed that loop retrieval can be largely improved by origin-shifting, coupled place similarity/odometry uncertainty search, and selecting the most likely loop constraint as proposed by our verification model. A high level of robustness was achieved by carefully verifying loop constraints based on
multiple sources of information, such as place similarity, consistency with odometry uncertainty, and alignment quality assessed after registration.
We evaluated TBV on \changed{three public datasets and demonstrated a substantial improvement to the state of the art, while generalizing to new environments,} making radar an attractive option to lidar for robust and accurate localization.
Some findings in our ablation study suggest that heavy filtering is undesired as it discards \changed{important details} for place recognition. Thus, in the future, we will explore building detailed and dense representations of scenes, fully utilizing the geometric information richness, uniquely provided by spinning FMCW radar.


\bibliographystyle{Journal/IEEEtran}
\bibliography{Journal/references}

\end{document}